\documentclass[final,onefignum,onetabnum]{siamart190516}
\pdfoutput=1 

\usepackage{upref}

\usepackage{amsfonts}
\usepackage{graphicx}
\usepackage{algorithmic}
\usepackage{amsopn}

\usepackage{enumitem}
\usepackage{amssymb} 
\usepackage{subcaption} 
\usepackage{booktabs} 
\usepackage[italicdiff]{physics} 
\usepackage{mathtools} 
\usepackage{bbm} 
\let\originalleft\left 
\let\originalright\right
\renewcommand{\left}{\mathopen{}\mathclose\bgroup\originalleft}
\renewcommand{\right}{\aftergroup\egroup\originalright}
\newcommand{\one}{\mathbbm{1}} 
\newcommand{\E}{\mathbb{E}}
\renewcommand{\P}{\mathbb{P}}
\newcommand{\N}{\mathbb{N}}
\newcommand{\Z}{\mathbb{Z}}
\newcommand{\R}{\mathbb{R}}
\newcommand{\cP}{\mathcal{P}}
\newcommand{\cA}{\mathcal{A}}
\newcommand{\cL}{\mathcal{L}}
\newcommand{\cM}{\mathcal{M}}
\newcommand{\cF}{\mathcal{F}}
\newcommand{\cH}{\mathcal{H}}
\newcommand{\cX}{\mathcal{X}}
\newcommand{\cY}{\mathcal{Y}}
\newcommand{\cD}{\mathcal{D}}
\newcommand{\cC}{\mathcal{C}}
\newcommand{\al}{\alpha}
\newcommand{\ep}{\varepsilon}
\DeclarePairedDelimiterX{\iptemp}[2]{\langle}{\rangle}{#1, #2}
\renewcommand{\ip}{\iptemp}
\DeclarePairedDelimiterX{\normtemp}[1]{\lVert}{\rVert}{#1}
\renewcommand{\norm}{\normtemp}
\DeclarePairedDelimiterX{\abstemp}[1]{\lvert}{\rvert}{#1}
\renewcommand{\abs}{\abstemp}
\newcommand{\ON}[1]{\operatorname{#1}}
\newcommand{\lap}{\Delta}
\DeclareMathOperator{\id}{Id} 
\DeclareMathOperator*{\argmin}{argmin}
\DeclareMathOperator{\im}{im} 

\newcommand{\Fd}{F^{\dagger}} 
\newcommand{\defeq}{\coloneqq} 
\newcommand{\diid}{\stackrel{\mathrm{iid}}{\sim}} 

\def\qfa{\quad\text{for all}\quad}

\def\qa{\quad\text{and}\quad}
\def\qf{\quad\text{for}\quad}
\def\qw{\quad\text{where}\quad}
\def\qin{\quad\text{in}\quad}
\def\qon{\quad\text{on}\quad}

\newcommand{\slot}{{\,\cdot\,}}

\ifpdf
\DeclareGraphicsExtensions{.eps,.pdf,.png,.jpg}
\else
\DeclareGraphicsExtensions{.eps}
\fi

\newsiamremark{remark}{Remark}
\newsiamremark{hypothesis}{Hypothesis}
\crefname{hypothesis}{Hypothesis}{Hypotheses}
\newsiamthm{claim}{Claim}
\newsiamremark{example}{Example}
\newsiamremark{notation}{Notation}
\newsiamthm{result}{Result}
\newsiamthm{assumption}{Assumption}
\crefname{assumption}{Assumption}{Assumptions}

\makeatletter
\long\def\@makecaption#1#2{%
    \footnotesize%
    \setlength{\parindent}{1.5pc}%
  \ifx\@captype\@figtxt
    \vskip\abovecaptionskip
    \setbox\@tempboxa\hbox{{\normalfont\scshape #1}. {\normalfont\itshape #2}}%
    \ifdim \wd\@tempboxa >\hsize
      {{\normalfont\scshape #1}. {\normalfont\itshape #2}\par}
    \else
      \global\@minipagefalse
      \centering\normalfont{\scshape #1}. {\itshape #2}\par
    \fi
  \else
    \setbox\@tempboxa\hbox{{\normalfont\itshape #2}}%
    \ifdim \wd\@tempboxa >\hsize
       {\centering\normalfont\scshape #1\par}
      {\normalfont\itshape #2\par}
    \else
    \global\@minipagefalse
      \centering{\normalfont\scshape #1\par}{\normalfont\itshape #2\par}
    \fi
    \vskip\belowcaptionskip
  \fi}
\makeatother

\title{Operator Learning Using Random Features:\\A Tool for Scientific Computing\thanks{Published electronically August 8, 2024 in the SIGEST section of \textit{SIAM Review}. The corresponding SIGEST editorial commentary may be found at the following link: \url{https://doi.org/10.1137/24N975943}. The present paper is an expanded version of an article that originally appeared in \textit{SIAM Journal on Scientific Computing}, Volume 43, Number 5, 2021, pages A3212--A3243, under the title ``The Random Feature Model for Input-Output Maps between Banach Spaces.''
\funding{The original work~\cite{nelsen2021random} was supported by the National Science Foundation (NSF) Graduate Research
Fellowship Program under award DGE-1745301, NSF award DMS-1818977, Office of Naval Research (ONR) award N00014-17-1-2079, NSF award AGS-1835860, and ONR award N00014-19-1-2408. This SIGEST article is supported by the Amazon/Caltech
AI4Science Fellowship held by the first author and by the Department of Defense
Vannevar Bush Faculty Fellowship, under ONR award N00014-22-1-2790, held by the second author.}}}

\author{Nicholas H. Nelsen\thanks{Department of Computing and Mathematical Sciences, California Institute of Technology, Pasadena, CA
91125 USA (\email{nnelsen@caltech.edu}, \email{astuart@caltech.edu}).}
\and Andrew M. Stuart{\footnotemark[2]}}

\headers{Operator Learning Using Random Features}{Nicholas H. Nelsen and Andrew M. Stuart}

\ifpdf
\hypersetup{
  pdftitle={Operator Learning Using Random Features: A Tool for Scientific Computing
},
  pdfauthor={N. H. Nelsen and A. M. Stuart}
}
\fi

\begin{document}

\maketitle

\begin{abstract}
Supervised operator learning centers on the use of training data, in the
form of input-output pairs, to estimate maps between infinite-dimensional spaces. It is emerging as a powerful tool to complement traditional scientific computing, which may often be framed in terms of operators mapping between spaces of functions. Building on the classical random features methodology for scalar regression, this paper introduces the function-valued random features method. This leads to a supervised operator learning architecture that is practical for nonlinear problems yet is structured enough to facilitate efficient training through the optimization of a convex, quadratic cost. Due to the quadratic structure, the trained model is equipped with convergence guarantees and error and complexity bounds, properties that are not readily available for most other operator learning architectures. At its core, the proposed approach builds a linear combination of random operators. This turns out to be a low-rank approximation of an operator-valued kernel ridge regression algorithm, and hence the method also has strong connections to Gaussian process regression. The paper designs function-valued random features that are tailored to the structure of two nonlinear operator learning benchmark problems arising from parametric partial differential equations. Numerical results demonstrate the scalability, discretization invariance, and transferability of the function-valued random features method.
\end{abstract}

\begin{keywords}
scientific machine learning, operator learning, random features, surrogate model, kernel ridge regression, parametric partial differential equation
\end{keywords}

\begin{AMS}
68T05, 65D40, 62J07, 62M45, 68W20, 35R60
\end{AMS}

\section{Introduction}
\label{sec:intro}
The increased use of machine learning for complex scientific tasks ranging from drug discovery to numerical weather prediction has led to the emergence of the new field of \emph{scientific machine learning}. Scientific machine learning blends modern artificial intelligence techniques with time-tested scientific computing methods in a principled manner to tackle challenging science and engineering problems, even those previously considered to be out of reach due to high-dimensionality or computational cost.
A common theme in these physical problems is that the data are typically modeled as infinite-dimensional quantities like velocity or pressure fields. Such continuum objects are
spatially and temporally varying functions that have intrinsic smoothness properties and long-range
correlations. 

Recognizing the need for new mathematical development of learning algorithms that are tailor-made for continuum problems, researchers established the \emph{operator learning} paradigm to build data-driven models that map between infinite-dimensional input and output spaces. An operator is an input-output relationship such that each input and corresponding output is infinite-dimensional. For example, the mapping from the current temperature in a room to the temperature one hour later is an operator. This is because temperature at a fixed time is a function characterized by its values at an uncountably-infinite number of spatial locations. More generally, one may consider the semigroup generated by a time-dependent partial differential equation (PDE) mapping the initial condition to the solution at a later time. A more concrete example is the mapping $\Fd\colon (a,f)\mapsto u$ from coefficient function $a$ and source term $f$ to solution function $u$ governed by the elliptic PDE $-\nabla\cdot(a\nabla u)=f$, equipped with appropriate boundary conditions. The paper returns to this example later on.

The paper focuses exclusively on supervised operator learning. This is
concerned with learning models to fit infinite-dimensional input-output pairs of (what is then known as labeled) training data. However, the operator learning framework is quite general and encompasses continuum problems that involve potentially diverse sources of data, going beyond the supervised learning setting. In the unsupervised setting, only unlabeled data is available. One example of this is the estimation of a covariance operator: the dataset comprises random functions drawn from the
probability measure whose covariance is to be estimated. Alternatively, the observed data might only consist of indirect or sparse measurements of a system, often also corrupted by noise, as is common in inverse problems; blind deconvolution is an important example.

Infinite-dimensional quantities must always be discretized when represented on a computer or in experiments. What distinguishes supervised operator learning from traditional supervised learning architectures that operate on high-dimensional discretized vectors is that in the continuum limit of infinite resolution, operator learning architectures have a well-defined and consistent meaning. They capture the underlying continuum structure of the problem and not artifacts due to the particular choice of discretization. Indeed, for a fixed set of trainable parameters, operator learning methods by construction produce consistent results given any finite-dimensional discretization of the conceptually infinite-dimensional data. That is, they are inherently dimension- and discretization-independent. Practically, this means that once learned at one resolution, the operator can be
transferred to any other resolution without the need for re-training. Growing empirical evidence suggests that operator learning exhibits excellent performance as a tool to accelerate model-centric tasks in science and engineering or to discover unknown physical laws from experimental data. However, the mathematical theory of operator learning is far from complete, and this limits its impact.

The goal of this paper is to develop an operator learning methodology with strong theoretical foundations that is also especially well-suited for the task of speeding up otherwise prohibitively expensive many-query problems. The need for repeated evaluations of a complex, costly, and slow forward model for different configurations of a system parameter occurs in various science and engineering domains. The true model is often a PDE and the parameter, serving as input to the PDE model, is often a continuum quantity. For instance, in the heat equation, the input is its initial condition, and in the preceding elliptic PDE example, the input is its coefficient and forcing functions. In contrast to the big data regime that dominates computer vision and other technological fields, only a relatively small amount of high resolution labeled data can be generated from computer simulations or physical experiments in scientific applications. Fast approximate surrogates built from this limited available data that can efficiently and accurately emulate the full order model would be highly advantageous in downstream outer loop applications.

The present work demonstrates that the \emph{random feature model} (RFM) has considerable potential for such a purpose when formulated as a map between function spaces. In contrast to more complicated deep learning approaches, the function-valued random features algorithm involves learning the coefficients of a linear expansion composed of random maps. For a suitable training objective function, this is a finite-dimensional convex, quadratic optimization problem. Equivalently, the paper shows that this supervised training procedure is equivalent to ridge regression over a reproducing kernel Hilbert space (RKHS) of operators. As a consequence of the careful construction of the method as mapping between infinite-dimensional Banach spaces, the resulting RFM surrogate enjoys rigorous convergence guarantees and scales favorably with respect to (w.r.t.) the high input and output dimensions arising in practical, discretized applications. Numerically, the method achieves small test error for learning a semigroup and the solution operator of a parametric elliptic PDE.

This section continues with a literature review and then a summary of the main contributions of the paper.

\subsection{Literature Review}\label{sec:intro_lit}
Two different lines of research have emerged that address PDE approximation problems with scientific machine learning techniques. The first perspective takes a more traditional approach akin to point collocation methods from the field of numerical analysis. Here, the goal is to use a deep neural network (NN)~\cite{raissi2019physics} or other function class~\cite{chen2022bridge} to solve a prescribed initial boundary value problem with as high accuracy as possible. Given a point cloud in a possibly high-dimensional spatio-temporal domain $ \cD $ as input data, the prevailing approach first directly parametrizes the PDE solution field as an NN and then optimizes the NN parameters by minimizing the PDE residual w.r.t. some loss functional using variants of stochastic gradient descent~(see~\cite{karniadakis2021physics,raissi2019physics,sirignano2018dgm,weinan2018deep} and the references therein). To clarify, the object approximated with this approach is a function $\cD \to\R $ between finite-dimensional spaces. While mesh-free by definition, the method is highly intrusive as it requires full knowledge of the specified PDE. Any change to the original formulation of the initial boundary value problem or related PDE problem parameters necessitates an expensive retraining of the NN approximate solution. We do not explore this first approach any further in this paper.

The second direction takes an operator learning perspective and is arguably more ambitious: use an NN to emulate the infinite-dimensional mapping between an input parameter and the PDE solution itself~\cite{boulle2023mathematical} or a functional of the solution, i.e., a quantity of interest~\cite{huang2024fnm}; the latter is widely prevalent in inverse problems~\cite{arridge2019solving,stuart2010inverse}, optimization under uncertainty~\cite{luo2023efficient}, or optimal experimental design~\cite{alexanderian2021optimal}. For an approximation-theoretic treatment of parametric PDEs, we mention the paper~\cite{cohen2015approximation}. We emphasize that the object approximated in this setting, unlike in the first approach mentioned in the previous paragraph, is an operator $ \cX\to\cY $, i.e., the PDE solution operator, where $ \cX $ and $ \cY $ are infinite-dimensional Banach spaces; this map is generally nonlinear. It is this second line of research that inspires our work. We now highlight several subtopics relevant to surrogate modeling and operator learning. Summaries of the state-of-the-art for operator learning may be found in two mathematically-oriented review articles on the subject~\cite{boulle2023mathematical,kovachki2024operator}.

\subsubsection*{Model Reduction}
There are many approaches to surrogate modeling that do not explicitly involve machine learning ideas~\cite{peherstorfer2018survey}. The reduced basis method~(see \cite{barrault2004empirical,benner2017model,devore2014theoretical} and the references therein) is a classical idea based on constructing an empirical basis from continuum or high-dimensional data snapshots and solving a cheaper variational problem; it is still widely used in practice due to computationally efficient offline-online decompositions that eliminate dependence on the full order degrees of freedom. Machine learning extensions to the reduced basis methodology, of both intrusive (e.g., projection-based reduced order models) and nonintrusive (e.g., model-free data only) type, have further improved the applicability of these methods~\cite{barnett2023neural,cheng2013data,gao2019non,geelen2023operator,hesthaven2018non,lee2020model,qian2020lift,santo2019data}. However, the input-output maps considered in these works involve high dimension in only one of the input or the output space, not both. A line of research aiming to more closely align model reduction with operator learning is the work on deep learning-based reduced order models (ROMs)~\cite{brivio2023error,franco2024practical,franco2023approximation,franco2023latent}; some of these studies also derive approximation guarantees for the ROMs. Other popular surrogate modeling techniques include Gaussian processes~\cite{williams2006gaussian}, polynomial chaos expansions~\cite{spanos1989stochastic}, and radial basis functions~\cite{wendland2004scattered}, yet these are only practically suitable for scalar-valued maps with input space of low to moderate dimension unless strong assumptions are placed on the problem. Classical numerical methods for PDEs may also represent the discretized forward model as a map $ \R^{K}\to\R^{K} $, where $K$ is the resolution, albeit implicitly in the form of a computer code~(e.g., finite element, finite difference, finite volume methods). However, the approximation error is sensitive to $ K $ and repeated evaluations of this forward model often become cost prohibitive due to poor scaling with input dimension $ K $.

\subsubsection*{Operator Learning}
Many earlier attempts to build cheap-to-evaluate surrogate models for PDEs display sensitivity to discretization.
There is a suite of work on data-driven discretizations of PDEs that allows for identification of the governing system~\cite{bar2019learning,bigoni2020data,long2017pde,patel2020physics,stevens2020finitenet,trask2019gmls}. However, we note that only the operators appearing in the underlying equation itself are approximated with these approaches, not the solution operator of the PDE; the focus in these works is mostly on model discovery rather than model acceleration. More in line with the theme of the present paper, architectures based on deep convolutional NNs have proven quite successful for learning elliptic PDE solution operators. For example, see \cite{franco2024practical,tripathy2018deep,winovich2019convpde,zhu2018bayesian}, which take an image-to-image regression approach. Other NNs have been used in similar elliptic problems for quantity of interest prediction~\cite{huang2024fnm,khoo2017solving}, error estimation~\cite{chen2020output}, or unsupervised learning~\cite{li2020variational}, and for parametric PDEs more generally~\cite{geist2020numerical,kutyniok2019theoretical,opschoor2020deep,schwab2019deep}.
Yet in most of the preceding approaches, the architectures and resulting error are dependent on the mesh resolution. To circumvent this issue, the surrogate map must be well defined on function space and independent of any finite-dimensional realization of the map that arises from discretization. This is not a new idea~(see~\cite{chen1995universal,mhaskar1997neural,rossi2005functional} or, for functional data analysis,~\cite{kadri2016operator,micchelli2005learning,ramsay1991some}). The aforementioned reduced basis method is an example, as is the method of~\cite{chkifa2013sparse, cohen2015approximation}, which approximates the solution map with sparse Taylor polynomials and achieves optimal convergence rates in idealized settings. Early work in the use of NNs to learn the solution
operator, or vector field, defining ODEs and time-dependent PDEs
may be found from the 1990s \cite{chen1995universal,Kev98,mhaskar1997neural,Kev92}. However, only recently have \emph{practical} machine learning methods been designed to directly operate in infinite dimensions. 

Several implementable operator learning architectures were developed concurrently \cite{adcock2020deep, bhattacharya2020pca, li2020neural, lu2019deeponet, nelsen2021random, o2020derivative, wu2020data}. These include the DeepONet~\cite{lu2019deeponet}, which generalizes and makes practical the main idea in~\cite{chen1995universal}, PCA-Net \cite{bhattacharya2020pca}, and the RFM from the original version of the present paper~\cite{nelsen2021random}. These were followed by neural operators~\cite{kovachki2023neural,li2020neural} and in particular the Fourier Neural Operator~\cite{li2020fourier}. Details for and comparisons among these architectures are given in~\cite[sect. 3]{kovachki2024operator}. Apart from the RFM, what these methods---which we collectively call ``neural operators''---share is a deep learning backbone. The approximation theory of such neural operators is fairly well developed~\cite{herrmann2022neural,huang2024fnm,korolev2021two,kovachki2021universal,kovachki2023neural, lanthaler2023operator,lanthaler2023nonlocal,lanthaler2022deep,lanthaler2023curse}. It includes qualitative universal approximation, i.e., density, results as well as quantitative parameter complexity bounds, that is, the number of NN parameters required to achieve accuracy $\ep$. The paper \cite{lanthaler2023curse} reveals a ``curse of parametric complexity'' in which the parameter complexity is shown to be exponentially large in powers of $\ep^{-1}$ to approximate general Lipschitz continuous operators. This aligns with the findings of older work~\cite{mhaskar1997neural} and suggests that efficient neural operator learning is not possible without further assumptions. It turns out that the curse is lifted if enough regularity is assumed. For example, for linear or holomorphic target operators, efficient algebraic approximation rates may be established~\cite{adcock2022near,herrmann2022neural}.
However, what rates are possible for sets of operators ``in between'' holomorphic and Lipschitz operators is still an open question.

One of the simplest classes of operators are linear ones. There is a substantial body of work in this setting ranging from the learning of general linear operators~\cite{de2023convergence,jin2022minimax,mollenhauer2022learning,stepaniants2023learning} to estimating the Green's function of specific linear PDEs~\cite{gin2021deepgreen,boulle2023elliptic,boulle2023learning,schafer2024sparse} and Koopman operators~\cite{kostic2022learning}.
The linear setting allows for very thorough and sharp statistical analysis that leads to deep insights about the data efficiency of operator learning in terms of problem structure~\cite{boulle2023elliptic,de2023convergence,huang2024fnm}. Some sample complexity results have been obtained for nonlinear functionals and operators, which give the training dataset size needed to obtain $\ep$ accuracy. Most of this theory depends on kernels, either in an RKHS framework~\cite{caponnetto2007optimal,lanthaler2023error} or via local averaging (e.g., kernel smoothers)~\cite{ferraty2007nonparametric,oliva2013distribution}. Error bounds are obtained for encoder--decoder neural operators such as DeepONet and PCA-Net in~\cite{liu2024deep}. These results imply a ``curse of sample complexity,'' i.e., exponentially large sample sizes, for learning Lipschitz operators. Similar to the parameter complexity case, with enough regularity assumed on the operators of interest, as expressed through weighted tensor product structure, operator holomorphy, or analyticity, for example, minimax lower bounds can return to better behaved algebraic rates in the sample size~\cite{adcock2024optimal,caponnetto2007optimal,ingster2011estimation,ingster2009estimation}. Moreover, there exist both constructive and non-constructive estimators that achieve these algebraic convergence rates for operator learning~\cite{adcock2022near,caponnetto2007optimal,lanthaler2023error}.

\subsubsection*{Random Features}
The RFM as a mapping between finite-dimensional spaces was formalized in the series of papers
\cite{rahimi2008random, rahimi2008uniform, rahimi2008weighted}, building on earlier precursors in~\cite{barron1993universal,neal1996priors,williams1997computing}. The RFM is in some sense the simplest possible machine learning model; it may be viewed as an ensemble average of randomly parametrized functions: an expansion in a randomized basis with trainable coefficients.
The method of random Fourier features is the most mainstream instantiation of the approach~\cite{rahimi2008random}. Here, the RFM is used to approximate popular translation-invariant kernels by averages of sinusoidal functions with random frequencies. This approximation is then used downstream for kernel regression tasks~\cite{hashemi2023generalization}. An equivalent viewpoint is that the RFM approximates the Gaussian process prior distribution in a Gaussian process regression method~\cite{williams2006gaussian}. However, the choice of random feature map can be much more general than just random sines and cosines. These random features could be defined, for example, by randomizing the internal parameters of an NN. Many papers take this viewpoint~\cite{gonon2023approximation,mei2022generalization,rahimi2008uniform,rahimi2008weighted}. Compared to NN emulators with enormous learnable parameter counts~(e.g., $ O(10^{5}) $ to $ O(10^7)$; see~\cite{fan2020solving,feliu2020meta,li2020variational}) and methods that are intrusive or lead to nontrivial implementations~\cite{chkifa2013sparse,lee2020model,santo2019data}, the RFM is one of the simplest models to formulate and train. Often $ O(10^{4})$ or fewer linear expansion coefficients---which are the only free parameters in the RFM---suffice. 

The theory of the RFM for real-valued outputs is well developed, partly due to its close connection to kernel methods~\cite{bach2017equivalence,cao2019generalization,jacot2018neural,rahimi2008random,wendland2004scattered} and Gaussian processes~\cite{neal1996priors,williams1997computing}, and includes generalization bounds and dimension-free rates~\cite{lanthaler2023error,li2021towards,ma2019generalization,rahimi2008uniform,rudi2017generalization,sun2018theory,sun2018random}. A quadrature viewpoint on the RFM provides further insight and leads to Monte Carlo sampling ideas~\cite{bach2017equivalence}. As in modern deep learning practice, for some problem classes the RFM has been shown to perform well even when overparametrized~\cite{belkin2019reconciling,ma2019generalization,mei2022generalization}. However, overparametrization is not necessary for good performance; state-of-the-art fast rates are established in the underparametrized regime by~\cite{li2021towards,rudi2017generalization}. The paper \cite{gonon2023approximation} derives similar bounds for random neural network approximation of functionals with a random feature-based training strategy.

For the supervised operator learning setting in which inputs and outputs are both infinite-dimensional, kernel~\cite{caponnetto2007optimal,kadri2016operator} and Gaussian process methods~\cite{batlle2024kernel}---and hence random features---are less explored.
The paper \cite{oliva2015fast} performs nonlinear operator learning in the encoder--decoder paradigm, where the input and output spaces are represented by truncated orthonormal bases and the finite-dimensional coefficient-to-coefficient mapping is performed with a kernel smoother. The kernel smoother is then approximated with random Fourier features~\cite{rahimi2008random}. A similar idea is undertaken from a Gaussian process perspective~\cite{batlle2024kernel}. For high-dimensional input parameter spaces, the authors of~\cite{griebel2017reproducing,kempf2017kernel} analyze nonparametric kernel regression for parametric PDEs with real-valued solution map outputs. However, the preceding methods have poor computational scalability w.r.t. data dimension and sample size. The RFM alleviates these issues with randomization and efficient convex optimization.
The specific random Fourier feature approach of Rahimi and Recht~\cite{rahimi2008random} was generalized in \cite{brault2016random} to the finite-dimensional matrix-valued kernel setting with vector-valued random Fourier features and to the operator-valued kernel setting in~\cite{minh2016operator}. However, canonical operator-valued kernels are hard to define and the preceding works require explicit knowledge of these kernels. Our viewpoint in the current paper is to develop function-valued random features and work directly with these as a standalone supervised learning method, choosing them for their properties and noting that they implicitly define a kernel, but not working directly with this kernel. An additional benefit of our approach is that it avoids the nonconvex training routines that plague more sophisticated neural operator architectures and, in particular, hinder the development of uncertainty quantification and comprehensive complexity bounds. The key idea underlying our methodology is to formulate the proposed operator random features algorithm on infinite-dimensional space and only discretize it at implementation time. This philosophy in algorithm development has been instructive in a number of areas in scientific computing, as we describe next.

\subsubsection*{Other Continuum Algorithms}
The general philosophy of designing algorithms at the continuum level has been hugely
successful across disciplines. In PDE-constrained optimization, there is the ``optimize-then-discretize'' principle~\cite{hinze2008optimization}. In applied probability, there are Markov chain Monte Carlo algorithms for sampling probability distributions supported on function spaces~\cite{cotter2013mcmc}. The Bayesian formulation of inverse problems on Banach spaces provides another example~\cite{stuart2010inverse}. There is work along similar lines that extends numerical linear algebra routines for finite-dimensional vectors and matrices to new ones for infinite-dimensional functions and linear operators~\cite{townsend2013extension,townsend2015continuous}. All such methods inherit certain dimension independent properties that make them more robust and possibly more accurate. Operator learning brings this powerful perspective to machine learning, where it has been promoted as a way of designing and analyzing learning algorithms \cite{haber2017stable,ma2019machine,ruthotto2019deep,weinan2017proposal,weinan2019mean}. Our work may be understood within this general framework.

\subsection{Contributions}\label{sec:intro_contrib}
Our primary contributions in this paper are now listed.
\begin{enumerate}[label=(C\arabic*)]
        \item We develop the RFM, directly formulated on the function space level, for learning nonlinear operators between Banach spaces purely from data. As a method for parametric PDEs, the methodology is non-intrusive but also has the additional advantage that it may be used in settings where only data is available and no model is known.
        \item We show that our proposed method is more computationally tractable to both train and evaluate than standard kernel methods in infinite dimensions. Furthermore, we show that the method is equivalent to kernel ridge regression performed in a finite-dimensional space spanned by random features and comes equipped with a full convergence theory.
        \item We apply our operator learning methodology to learn the semigroup defined by the
        solution operator for the viscous Burgers' equation and the coefficient-to-solution operator for the Darcy flow equation.
        \item We perform numerical experiments that demonstrate two mesh-independent approximation properties that are built into the proposed methodology: invariance of relative error to mesh resolution and evaluation ability on any mesh resolution.
\end{enumerate}

The remainder of this paper is structured as follows. In~\cref{sec:problem}, we communicate the mathematical framework required to work with the RFM in infinite dimensions, identify an appropriate approximation space, explain the training procedure, and review recent error bounds for the method. We introduce two instantiations of random feature maps that target physical science applications in~\cref{sec:application} and detail the
corresponding numerical results for these applications in~\cref{sec:experiment}. We conclude in \cref{sec:conclusion} with a summary and directions future work.

\section{Methodology}
\label{sec:problem}
In this work, the overarching problem of interest is the approximation of a map $ \Fd\colon \cX\to\cY $, where $ \cX$ and $\cY $ are infinite-dimensional spaces of real-valued functions defined on some bounded open subset of $ \R^{d} $, and $\Fd $ is defined by $ a\mapsto \Fd(a)\defeq u $, where $ u\in\cY $ is the solution of a (possibly time-dependent) PDE and $a\in\cX$ is an input function required to make the problem well-posed. Our proposed approach for this approximation, constructing a surrogate map $ F $ for the true map $ \Fd $, is
data-driven, nonintrusive, and based on least squares. Least squares--based methods are integral to the random feature methodology as proposed in low dimensions~\cite{rahimi2008random,rahimi2008uniform} and generalized here to the infinite-dimensional setting. They have also been shown to work well in other algorithms for high-dimensional numerical approximation~\cite{beylkin2005algorithms,cohen2016optimal,doostan2009least}. Within the broader scope of reduced order modeling techniques~\cite{benner2017model}, the approach we adopt in this paper falls within the class of data-fit emulators. In its essence, our method approximates the solution manifold
\begin{align}\label{eqn:manifold}
\cM=\bigl\{u\in\cY\colon u=\Fd(a) \qa  a\in\cX\bigr\}
\end{align}
on average. The solution map $ \Fd $, often being the inverse of a differential operator, is usually smoothing and admits some notion of compactness. Then, the idea is that $ \cM $ should have some compact, low-dimensional structure or intrinsic dimension. However, actually finding a model $ F $ that exploits this structure despite the high dimensionality of the truth map $ \Fd $ is quite difficult. Further, the effectiveness of many model reduction techniques, such as those based on the reduced basis method, are dependent on inherent properties of the map $ \Fd $ itself (e.g., analyticity), which in turn may influence the decay rate of the Kolmogorov width of the manifold $ \cM $~\cite{cohen2015approximation}. While such subtleties of approximation theory are crucial to
developing rigorous theory and provably convergent algorithms~\cite{kovachki2024operator}, we choose to work in the nonintrusive setting where knowledge of the map $ \Fd $ and its associated PDE are only obtained through measurement data, and hence detailed characterizations such as those aforementioned are essentially unavailable. Thus, we emphasize that our proposed operator learning methodology is applicable to general continuum problems with function space data, not just to PDEs.

The remainder of this section introduces the mathematical preliminaries for our methodology. With the goal of operator approximation in mind, in~\cref{sec:problem_form_SL} we formulate a supervised learning problem in an infinite-dimensional setting. We provide the necessary background on RKHSs in~\cref{sec:rkhs} and then define the RFM in~\cref{sec:rfm}. In~\cref{sec:opt}, we describe the optimization principle which leads to implementable algorithms for the RFM and an example problem in which $\cX$ and $\cY$ are one-dimensional vector spaces. We finish by providing two convergence results for trained function-valued RFMs in \cref{sec:theory}.

\subsection{Problem Formulation}
\label{sec:problem_form_SL}
Let $ \cX$ and $ \cY $ be real Banach spaces and $ \Fd\colon\cX\to\cY $ be a (possibly nonlinear) map. It is natural to frame the approximation of $ \Fd $ as a supervised learning problem. Suppose we are given training data in the form of input-output pairs $ \{(a_i, y_i)\}_{i=1}^{n}\subset \cX\times\cY$, where $ a_i\sim \nu$ are independent and identically distributed (i.i.d.), $ \nu $ is a probability measure supported on $ \cX $, and $ y_i$ is given by $\Fd(a_i) \sim \Fd_{\sharp}\nu$ plus, potentially, noise. In the examples in this paper, the noise is viewed  as resulting from model error (the PDE does not perfectly represent the physics) or from discretization error (in approximating the PDE); situations in which the data acquisition process is inherently noisy can also be envisioned~\cite{lanthaler2023error} but are not explicitly studied here. We aim to build a parametric reconstruction of the true map $ \Fd $ from the data, that is, construct a model $ F\colon\cX\times\cP\to\cY $ and find $ \al^{\dagger}\in\cP\subseteq\R^{m} $ such that $ F(\slot, \al^{\dagger})\approx \Fd $ are close as maps from $\cX$ to $\cY$ in some suitable sense. The natural number $ m $ here denotes the total number of model parameters.

The standard approach to determine parameters in supervised learning is to define a loss functional $ \ell\colon\cY\times\cY\to \R_{\geq 0} $ and minimize the expected risk,
\begin{align}\label{eqn:risk_expected}
\min_{\al\in\cP}\E^{a\sim\nu}\Bigl[\ell\bigl(\Fd(a), F(a,\al)\bigr)\Bigr]\, .
\end{align}
With only the data $ \{(a_i, y_i)\}_{i=1}^{n} $ at our disposal, we approximate problem~\cref{eqn:risk_expected} by replacing $ \nu $ with the empirical measure $ \nu^{(n)}\defeq\frac{1}{n}\sum_{i=1}^{n}\delta_{a_{i}} $, which leads to the empirical risk minimization problem
\begin{align}\label{eqn:risk_empirical}
\min_{\al\in\cP}\dfrac{1}{n}\sum_{i=1}^{n}\ell\bigl(y_i, F(a_i,\al)\bigr)\, .
\end{align}
The hope is that given minimizer $ \al^{(n)} $ of~\cref{eqn:risk_empirical} and $ \al^{\dagger} $ of~\cref{eqn:risk_expected}, $ F(\slot,\al^{(n)}) $ well approximates $ F(\slot, \al^{\dagger}) $, that is, the learned model \emph{generalizes} well; these ideas may be made rigorous with results from statistical learning theory~\cite{hastie2009elements}. Solving problem~\cref{eqn:risk_empirical} is called \emph{training} the model $ F $. Once trained, the model is then validated on a new set of i.i.d. input-output pairs previously unseen during the training process. This \emph{testing} phase indicates how well $ F $ approximates $ \Fd $. From here on out, we assume that $ (\cY, \ip{\cdot}{\cdot}_{\cY}, \norm{\cdot}_{\cY}) $ is a real separable Hilbert space and focus on the squared loss
\begin{align}\label{eqn:loss_square}
\ell(y,y')\defeq \dfrac{1}{2}\norm*{y-y'}^{2}_{\cY}\, .
\end{align}
We stress that our entire formulation is in an infinite-dimensional setting and we will remain in this setting throughout the paper; as such, the random feature methodology we propose will inherit desirable discretization-invariant properties, to be observed in the numerical experiments of~\cref{sec:experiment}.
\begin{notation}[expectation]
        For a Borel measurable map $ G\colon\mathcal{U}\to\mathcal{V} $ between two Banach spaces $ \mathcal{U}$ and $\mathcal{V} $ and a probability measure $ \pi $ supported on $ \mathcal{U} $, we denote the expectation of $ G $ under $ \pi $ by
        \begin{align}\label{eqn:expect_nton}
        \E^{u\sim \pi}\bigl[G(u)\bigr]=\int_{\mathcal{U}} G(u)\pi(du)\in\mathcal{V}
        \end{align}
        in the sense of Bochner integration (see, e.g., \cite[sect.~A.2]{Dashti2017}).
\end{notation}

\subsection{Operator-Valued Reproducing Kernels}\label{sec:rkhs}
The RFM is naturally formulated in an RKHS setting, as our exposition will demonstrate in~\cref{sec:rfm}. However, the usual RKHS theory is concerned with real-valued functions~\cite{aronszajn1950theory,berlinet2011reproducing,cucker2002mathematical,wendland2004scattered}. Our setting, with the output space $ \cY $ a separable Hilbert space, requires several ideas that generalize the real-valued case. We now outline these ideas with a review of operator-valued kernels; parts of the presentation that follow may be found in the references~\cite{bach2017equivalence,carmeli2006vector,micchelli2005learning,nelsen2022ovk}.

We first consider the special case $ \cY\defeq \R $ for ease of exposition. A real RKHS is a Hilbert space $ (\cH, \ip{\cdot}{\cdot}_{\cH}, \norm{\cdot}_{\cH}) $ comprising real-valued functions $ f\colon\cX\to\R $ such that the pointwise evaluation functional $ f\mapsto f(a) $ is bounded for every $ a\in\cX $. It then follows that there exists a unique, symmetric, positive definite kernel function $ k\colon \cX\times\cX\to\R $ such that for every $ a\in\cX $, we have $ k(\cdot, a)\in\cH $ and the \emph{reproducing kernel property} $ f(a)=\ip{k(\cdot,a)}{f}_{\cH} $ holds. These two properties are often taken as the definition of an RKHS. The converse direction is also true: every symmetric, positive definite kernel defines a unique RKHS~\cite{aronszajn1950theory}.

We now introduce the needed generalization of the reproducing property to the case of arbitrary real Hilbert spaces $ \cY $, as this result will motivate the construction of the RFM. Kernels in this setting are now operator-valued.
\begin{definition}[operator-valued kernel]\label{def:kernel_operatorvalued}
        Let $ \cX $ be a real Banach space and $ \cY $ a real separable Hilbert space. An \emph{{operator-valued kernel}} is a map
        \begin{align}\label{eqn:kernel_operatorvalued}
        k\colon\cX\times \cX\to\cL(\cY)\, ,
        \end{align}
        where $ \cL(\cY) $ denotes the Banach space of all bounded linear operators on $ \cY $, such that its adjoint satisfies $ k(a,a')^{*}=k(a',a) $ for all $ a$ and $ a' $ in $\cX $ and for every $ N\in\N $,
        \begin{align}\label{eqn:kernel_psd}
        \sum_{i=1}^{N}\sum_{j=1}^{N}\ip[\big]{y_i}{k(a_i,a_j)y_j}_{\cY}\geq 0
        \end{align}
        for all pairs $ \{(a_i, y_i)\}_{i=1}^{N}\subset \cX\times\cY$.
\end{definition}

Paralleling the development for the real-valued case, an operator-valued kernel $ k $ also uniquely (up to isomorphism) determines an associated real RKHS $ \cH_{k}=\cH_{k}(\cX;\cY) $ of operators mapping $\cX$ to $\cY$. Now, choosing a probability measure $ \nu $ supported on $ \cX $, we define a kernel integral operator $ T_k $ associated to $ k $ by
\begin{align}\label{eqn:integral_operator}
\begin{split}
T_{k}\colon L^{2}_{\nu}(\cX;\cY)&\to L^2_{\nu}(\cX;\cY)\, ,\\
F &\mapsto T_{k}F\defeq \E^{a' \sim \nu} \bigl[k(\cdot,a')F(a')\bigr]\, ,
\end{split}
\end{align}
which is nonnegative, self-adjoint, and compact (provided $ k(a,a)\in\cL(\cY) $ is compact for all $ a\in\cX $~\cite{carmeli2006vector}). Let us further assume that all conditions needed  for $ T_{k}^{1/2} $ to be an isometry from $ L^{2}_{\nu} $ into $ \cH_{k} $ are satisfied, i.e., $ \cH_{k}=\im(T_k^{1/2}) $. Generalizing the standard Mercer theory~(see, e.g.,~\cite{bach2017equivalence,berlinet2011reproducing}), we may write the RKHS inner product as
\begin{align}\label{eqn:ip_rkhs}
\ip{F}{G}_{\cH_{k}} = \ip{F}{T_{k}^{-1}G}_{L^2_{\nu}} \qfa F\in\cH_k \qa G\in\cH_{k}\, .
\end{align}
Note that while~\cref{eqn:ip_rkhs} appears to depend on the measure $ \nu $ on $ \cX $, the set $ \cH_{k} $ itself is determined by the kernel without any reference to a measure~\cite[Chap. 3, Theorem 4]{cucker2002mathematical}. With the inner product now explicit, it is possible to deduce the following reproducing property of the operator-valued kernel $k$~\cite[sect. 2.2]{nelsen2021random}.

\begin{result}[reproducing property for operator-valued kernels]
    Let $ F\in\cH_{k} $ be given. For every $ a\in\cX $ and $ y\in\cY $, it holds that
    \begin{align}\label{eqn:rkprop_banach}
    \ip{y}{F(a)}_{\cY}=\ip{k(\cdot,a)y}{F}_{\cH_{k}}\, .
    \end{align}
\end{result}

The identity \cref{eqn:rkprop_banach}, paired with a special choice of operator-valued kernel $ k $, is the basis of the RFM in our abstract infinite-dimensional setting.

\subsection{Random Feature Model}\label{sec:rfm}
One could approach the approximation of target map $ \Fd\colon\cX\to\cY $ from the perspective of kernel methods. However, it is generally a difficult task to explicitly design operator-valued kernels of the form~\cref{eqn:kernel_operatorvalued} since the spaces $ \cX $ and $\cY $ may be of different regularity, for example. Example constructions of operator-valued kernels studied in the literature include those taking value as diagonal operators, multiplication operators, or composition operators~\cite{kadri2016operator,micchelli2005learning,owhadi2022ideas}, but these all involve some simple generalization of scalar-valued kernels or strong assumptions about $\cY$. Instead, the RFM allows one to implicitly work with fully general operator-valued kernels through the use of a \emph{random feature map} $ \varphi\colon\cX\times\Theta\to\cY $ and a probability measure $ \mu $ supported on Banach space $ \Theta $. The map $\varphi$ is assumed to be square
integrable w.r.t.~the product measure $\nu \times \mu$, i.e., $ \varphi\in L_{\nu\times\mu}^2(\cX\times\Theta;\cY) $, where $ \nu $ is the (sometimes a modeling choice at our discretion, sometimes unknown) data distribution on $ \cX $. Together, $ (\varphi, \mu) $ form a \emph{random feature pair}. With this setup in place, we now describe the connection
between random features and kernels. To this end, recall the following standard notation.

\begin{notation}[outer product]
    Given a Hilbert space $ (H, \ip{\cdot}{\cdot}, \norm{\slot}) $, the \emph{outer product} $ a\otimes b\in\cL(H) $ is defined by $ (a\otimes b)c=\ip{b}{c} a $ for any $ a$, $b$, and $c\in H $.
\end{notation}

\subsubsection{An Intractable Nonparametric Model Class}\label{sec:rfm_infinite}
Given the pair $ (\varphi,\mu) $, we begin by considering maps $ k_{\mu}\colon\cX\times\cX\to\cL(\cY) $ of the form
\begin{align}\label{eqn:kernel_expectation}
k_{\mu}(a,a')\defeq \E^{\theta\sim\mu}\bigl[\varphi(a;\theta)\otimes \varphi(a';\theta)\bigr]\, .
\end{align}
Such representations need not be unique; different pairs $ (\varphi,\mu) $ may induce the same kernel $ k=k_{\mu} $ in \cref{eqn:kernel_expectation}. Since $ k_{\mu} $ may readily be shown to be an operator-valued kernel via~\cref{def:kernel_operatorvalued}, it defines a unique real RKHS $ \cH_{k_{\mu}}\subset L^{2}_{\nu}(\cX;\cY) $. Our methodology will be based on this space and, in particular, finite-dimensional approximations thereof.

We now perform a purely formal but instructive calculation, following
from application of the reproducing property~\cref{eqn:rkprop_banach}
to operator-valued kernels of the form~\cref{eqn:kernel_expectation}.
Doing so leads to an integral representation of any $ F\in\cH_{k_{\mu}}$. For all $ a\in\cX $ and $ y\in\cY $, it holds that
\begin{align*}
\ip{y}{F(a)}_{\cY}=\ip{k_{\mu}(\cdot, a)y}{F}_{\cH_{k_{\mu}}}
&=\ip[\Big]{\E^{\theta\sim\mu}\bigl[\ip{\varphi(a;\theta)}{y}_{\cY}\, \varphi(\slot;\theta)\bigr]}{F}_{\cH_{k_{\mu}}}\\
&=\E^{\theta\sim\mu}\Bigl[\ip{\varphi(a;\theta)}{y}_{\cY}\ip{\varphi(\slot;\theta)}{F}_{\cH_{k_{\mu}}}\Bigr]\\
&=\E^{\theta\sim\mu}\bigl[c_{F}(\theta)\ip{y}{\varphi(a;\theta)}_{\cY}\bigr]\\
&=\ip[\Big]{y}{\E^{\theta\sim\mu}\bigl[c_{F}(\theta)\varphi(a;\theta)\bigr]}_{\cY}\, ,
\end{align*}
where the coefficient function $ c_{F}\colon \Theta\to\R $ is defined by
\begin{align}\label{eqn:rfm_coeff_func}
\theta\mapsto c_{F}(\theta)\defeq \ip{\varphi(\slot;\theta)}{F}_{\cH_{k_{\mu}}}\, .
\end{align}
Since $ \cY $ is Hilbert, the above holding for all $ y\in\cY$
implies the integral representation
\begin{align}\label{eqn:rfm_integral_trans}
F = \E^{\theta\sim\mu}\bigl[c_{F}(\theta)\varphi(\slot;\theta)\bigr]\, .
\end{align}

The formal expression \cref{eqn:rfm_coeff_func} for $c_F(\theta)$ needs careful interpretation, which is provided in \cite[Appendix B]{nelsen2021random} of the original version of this paper. For instance, if $\varphi(\slot;\theta)$ is chosen to be a realization of a Gaussian process (as seen later in \Cref{ex:bb}), then $ \varphi(\slot;\theta) \notin \cH_{k_{\mu}}$ with
probability one; indeed, in this case $c_F$ is defined only as an
$L^2_{\mu}(\Theta;\R)$ limit. Nonetheless, the RKHS may be completely
characterized by this integral representation. Define the map
\begin{align}\label{eqn:rf_integral_operator}
\begin{split}
\cA\colon L^{2}_{\mu}(\Theta;\R)&\to L^2_{\nu}(\cX;\cY)\, ,\\
c &\mapsto \cA c\defeq \E^{\theta\sim\mu}\bigl[c(\theta)\varphi(\slot;\theta)\bigr] \, .
\end{split}
\end{align}
The map $ \cA $ may be shown to be a bounded linear operator that is a particular square root of $ T_{k_{\mu}} $ from \cref{eqn:integral_operator} \cite[Appendix B]{nelsen2021random}. We have the following result whose proof, provided in~\cite[Appendix A]{nelsen2021random} of the original version of this paper, is a straightforward
generalization of the real-valued case given in \cite[sect.~2.2]{bach2017equivalence}.

\begin{result}[infinite-dimensional RKHS]
\label{r:1}
    Under the assumption that the feature map $\varphi$ satisfies $\varphi \in L^2_{\nu \times \mu}(\cX \times \Theta; \cY)$, the RKHS defined by the kernel $ k_{\mu} $ in~\cref{eqn:kernel_expectation} is
    \begin{align}\label{eqn:rkhs_image}
    \cH_{k_{\mu}}=\im(\cA)=\biggl\{\E^{\theta\sim\mu}\bigl[c(\theta)\varphi(\slot;\theta)\bigr]\colon c\in L^{2}_{\mu}(\Theta;\R)\biggr\}\, .
    \end{align}
\end{result}

We stress that the integral representation of mappings in RKHS \cref{eqn:rkhs_image} is not unique since $ \cA $ is not injective in general. However, the particular choice $ c=c_{F} $ \cref{eqn:rfm_coeff_func} in representation \cref{eqn:rfm_integral_trans} does enjoy a sense of uniqueness as described in \cite[Appendix B]{nelsen2021random}. In particular, the $L^2_\mu(\Theta;\R)$ norm of $c_F$ equals the $\cH_{k_\mu}$ norm of $F$. The formula \cref{eqn:rkhs_image} suggests that $\cH_{k_\mu}$, which is built from $(\varphi,\mu)$ and completely determined by coefficient functionals $c\in L^2_\mu(\Theta;\R)$, is a natural nonparametric class of operators to perform approximation with. However, the actual implementation of estimators based on the model class $\cH_{k_\mu}$  is known to incur an enormous computational cost without further assumptions on the structure of $(\varphi,\mu)$, as we discuss later in this section. Instead, we next adopt a parametric approximation to this full RKHS approach.

\subsubsection{A Tractable Parametric Model Class}\label{sec:rfm_finite}
A central role in what follows is the approximation of measure $\mu$
by the empirical measure
\begin{align}
\label{eqn:em}
\mu^{(m)} \defeq  \frac{1}{m}\sum_{j=1}^{m}\delta_{\theta_{j}}\, ,\qw  \theta_{j}\diid\mu\, .
\end{align}
Given \cref{eqn:em}, define $ k^{(m)}\defeq k_{\mu^{(m)}} $ to be the empirical
approximation to $ k_{\mu} $, that is,
\begin{align}\label{eqn:kernel_empirical}
k^{(m)}(a,a')=\E^{\theta\sim\mu^{(m)}}\bigl[\varphi(a;\theta)\otimes\varphi(a';\theta)\bigr]=\dfrac{1}{m}\sum_{j=1}^{m}\varphi(a;\theta_{j})\otimes\varphi(a';\theta_{j})\, .
\end{align}
We then let $\cH_{k^{(m)}}$ be the unique RKHS induced by the
kernel $k^{(m)}$; note that $ k^{(m)} $ and hence $ \cH_{k^{(m)}} $ are themselves random. The following characterization of $\cH_{k^{(m)}}$
is proved in the original version of this paper~\cite[Appendix A]{nelsen2021random}.

\begin{result}[finite-dimensional RKHS]
\label{r:2}
Assume that $\varphi \in L^2_{\nu \times \mu}(\cX \times \Theta; \cY)$
and that the random features $\{\varphi(\slot;\theta_j)\}_{j=1}^m$ are linearly independent in $L^2_{\nu}(\cX;\cY)$. Then, the RKHS $\cH_{k^{(m)}}$ is equal to the linear span of $\{\varphi(\slot;\theta_j)\}_{j=1}^m$.
\end{result}

Applying a simple Monte Carlo sampling approach to elements in RKHS \cref{eqn:rkhs_image} by replacing probability measure $ \mu $ by empirical measure $\mu^{(m)}$ gives the intuition that
\begin{align}\label{eqn:MCRFM}
\frac{1}{m}\sum_{j=1}^{m}c(\theta_j)\varphi(\slot;\theta_j)\approx \E^{\theta\sim\mu}\bigl[c(\theta)\varphi(\slot;\theta)\bigr] \qf c\in L^2_{\mu}(\Theta;\R) \, .
\end{align}
This low-rank approximation achieves the Monte Carlo rate $ O(m^{-1/2})$ in expectation and, by virtue of \cref{r:2}, is in $\cH_{k^{(m)}}$. However, in the setting
of this work, the Monte Carlo approach does not give rise to
a practical method for learning a target map $ \Fd\in\cH_{k_{\mu}} $ because $\Fd$, $ k_{\mu} $, and $ \cH_{k_{\mu}} $ are all unknown; only the random feature pair $ (\varphi, \mu) $ is assumed to be given. Hence one cannot apply \cref{eqn:rfm_coeff_func} or \cite[equation (B.2), p. A3239]{nelsen2021random} to evaluate $ c=c_{\Fd} $ in \cref{eqn:MCRFM}. Furthermore, in realistic settings it may be that $ \Fd\not\in \cH_{k_{\mu}} $, which leads to an additional smoothness misspecification gap not accounted for by the Monte Carlo method. To sidestep these
difficulties, the RFM adopts a data-driven optimization approach
to determine a different estimator of $\Fd$, also from
the space $\cH_{k^{(m)}}$. We now define the RFM.

\begin{definition}[RFM]\label{def:rfm}
Given probability spaces $(\cX,\mathcal{B}(\cX),\nu)$ and $(\Theta,\mathcal{B}(\Theta), \mu)$ with $\cX$ and $ \Theta $ being real finite- or infinite-dimensional Banach spaces, a
real separable Hilbert space $\cY$, and
$\varphi \in L^2_{\nu \times \mu}(\cX \times \Theta; \cY)$,
the {\emph{RFM}} is the parametric map
\begin{align}\label{eqn:rfm}
\begin{split}
F_{m}\colon \cX\times\R^{m}&\to \cY\,,\\
(a; \al)&\mapsto F_{m}(a; \al)\defeq \dfrac{1}{m}\sum_{j=1}^{m}\al_{j}\varphi(a;\theta_{j})\, ,\qw \theta_{j}\diid\mu\, .
\end{split}
\end{align}
\end{definition}

We use the Borel $\sigma$-algebras $ \mathcal{B}(\cX)$ and $\mathcal{B}(\Theta) $ to define the probability spaces in the preceding definition. Our goal with the RFM is to choose parameters $\al\in\R^{m}$ so as to approximate mappings $ \Fd\in\cH_{k_{\mu}} $ (in the well-specified setting) by mappings $ F_{m}(\cdot;\al)\in\cH_{k^{(m)}} $. The RFM is itself random and may be viewed as a \emph{spectral method} because the randomized family $ \{\varphi(\slot;\theta)\} $ in the linear expansion~\cref{eqn:rfm} is defined $\nu$-almost everywhere on $ \cX$. 
Determining the coefficient vector $\al$ from data obviates the difficulties associated with the oracle Monte Carlo approach because the data-driven method only requires knowledge of the pair $ (\varphi,\mu) $ and knowledge of sample input-output pairs from target operator $\Fd$.

As written,~\cref{eqn:rfm} is incredibly simple. The operator $F_m$ is nonlinear in its input $a$ but linear in its coefficient parameters $\al$. In practice, the linearity w.r.t. the RFM parameters is broken by also learning \emph{hyperparameters} that appear in the pair $(\varphi,\mu)$. Moreover, similar to operator learning architectures such as neural operators~\cite{kovachki2023neural} and Fourier neural operators~\cite{li2020fourier}, the RFM is a \emph{nonlinear approximation}. This means that the output $F_m(a;\al)$ of the RFM belongs to a nonlinear manifold in $\cY$ (cp. equation \ref{eqn:manifold}) instead of a fixed linear subspace of $\cY$. In contrast, methods such as PCA-Net~\cite{bhattacharya2020pca} and DeepONet~\cite{lu2019deeponet} are restricted to such fixed linear spaces, which may limit their approximation power for specific classes of problems. More theory is required to quantitatively separate these two classes of approximation methods.

Overall, it is clear that the choice of random feature map and measure pair $ (\varphi, \mu) $ will determine the quality of approximation. Most papers deploying these methods, including \cite{brault2016random,rahimi2008random,rahimi2008uniform}, take a kernel-oriented perspective by first choosing a kernel and then finding a random feature map to estimate this kernel. Our perspective, more aligned with \cite{rahimi2008weighted,sun2018random}, is the opposite in that we allow the choice of random feature map $ \varphi $ and distribution $\mu$ to implicitly \emph{define} the kernel via the formula~\cref{eqn:kernel_expectation} instead of picking the kernel first. This viewpoint also has implications for numerics: the kernel never explicitly appears in any computations, which leads to memory and other cost savings. It does, however, leave
open the question of characterizing the universality~\cite{sun2018random} of such kernels and the RKHS $\cH_{k_{\mu}}$ of mappings from $\cX$ to $\cY$ that underlies the approximation method; this is an important avenue for future work.

\subsubsection{Connection to Neural Networks and Neural Operators}\label{sec:rfm_nn}
The close connection to kernels explains the origins of the RFM in the machine learning literature. Moreover, the RFM may also be interpreted in the context of NNs \cite{neal1996priors,sun2018random,williams1997computing,yehudai2019power}. To see this explicitly, consider the setting where $\cX$ and $\cY$ are both equal to the Euclidean space $\R$ and choose $ \varphi $ to be a family of hidden neurons of the form $ \varphi_{\text{NN}}(a;\theta)\defeq \sigma(\theta^{(1)}\cdot a + \theta^{(2)}) $, where $\sigma(\slot)$ is a nonlinear activation function. A single hidden layer NN would seek to find $\{(\alpha_j,\theta_j)\}_{j=1}^m \subset \R \times \R^{2}$ such that
\begin{align}
\label{eqn:vnn}
\frac{1}{m} \sum_{j=1}^m \alpha_j \varphi_{\text{NN}}(\slot;\theta_j)
\end{align}
matches the given training data $ \{(a_i, y_i)\}_{i=1}^{n}\subset \cX\times\cY$. More generally, and in arbitrary Euclidean spaces, one may allow
$ \varphi_{\text{NN}}(\slot;\theta)$ to be any deep NN.
However, while the RFM has the same \emph{form} as \cref{eqn:vnn}, there is a difference in the \emph{training}: the $ \theta_j $ are drawn i.i.d. from a probability measure and then fixed, and only the $\alpha_j$ are chosen to fit the training data. This idea immediately transfers to the operator learning setting in which $\cX$ and $\cY$ are function spaces and the maps $\varphi_{\text{NN}}(\slot;\theta)$ are themselves randomly initialized deep neural operators or DeepONets.
Given any deep NN with randomly initialized parameters, studies of the lazy training regime and neural tangent kernel \cite{cao2019generalization,jacot2018neural} suggest that adopting an RFM approach and only optimizing over the last layer weights $\al$ is quite natural. Indeed, it is observed that in this regime the internal NN parameters do not stray far from their random initialization during gradient descent while the last layer of parameters $\{\al_{j}\}_{j=1}^{m}$ adapt considerably.

Once the feature parameters $ \{\theta_{j}\}_{j=1}^{m}$ are sampled at random and fixed, training the RFM $ F_{m} $ only requires optimizing over $ \al\in\R^{m} $. Due to the linearity of $F_m$ in $\al$, this is a straightforward task that we now describe.

\subsection{Optimization}\label{sec:opt}
One of the most attractive characteristics of the RFM is its training procedure. With the $ L^2 $-type loss~\cref{eqn:loss_square} as in standard regression settings, optimizing the coefficients of the RFM w.r.t. the empirical risk \cref{eqn:risk_empirical} is a convex optimization problem, requiring only the solution of a finite-dimensional system of linear equations; the convexity also suggests the possibility of appending convex constraints (such as linear inequalities), although we do not pursue this here. Further, the kernels $ k_{\mu} $ or $ k^{(m)} $ are not required anywhere in the algorithm. We emphasize the simplicity of the underlying optimization tasks as they suggest the possibility of numerical implementation of the RFM into complicated black-box computer codes. This is in contrast with other methods such as deep neural operators, which are trained with variants of stochastic gradient descent. Such a training strategy leads to nonconvexity that is notoriously difficult to study both computationally and theoretically.

We now proceed to show that a regularized version of the quadratic optimization
problem~\cref{eqn:risk_empirical}--\cref{eqn:loss_square} arises naturally from approximation of a nonparametric regression problem defined over the RKHS $\cH_{k_{\mu}}.$ To this end, recall the supervised learning formulation in~\cref{sec:problem_form_SL}. Given $ n $ i.i.d. input-output pairs $ \{(a_i, y_i)\}_{i=1}^{n}\subset \cX\times\cY$ as data, with the $a_i$ drawn from (possibly unknown)
probability measure $\nu$ on $ \cX $ and $y_i=\Fd(a_i)$, the objective is to find an approximation $ \widehat{F} $ to the map $ \Fd $. Let $ \cH_{k_{\mu}} $ be the hypothesis space and $ k_{\mu} $ its operator-valued reproducing kernel of the form~\cref{eqn:kernel_expectation}. The most straightforward learning algorithm in this RKHS setting is kernel ridge regression, also known as penalized least squares. This method produces a nonparametric model by finding a minimizer $ \widehat{F} $ of
\begin{align}\label{eqn:general_leastsquares}
\min_{F\in\cH_{k_{\mu}}}\left\{ \sum_{i=1}^{n}\dfrac{1}{2}\norm[\big]{y_i-F(a_i)}_{\cY}^{2}+\dfrac{\lambda}{2}\norm[\big]{F}_{\cH_{k_{\mu}}}^{2} \right\}\, ,
\end{align}
where $ \lambda\geq 0 $ is a penalty parameter. By the representer theorem for operator-valued kernels \cite[Theorems 2 and 4]{micchelli2005learning}, the minimizer has the form
\begin{align}\label{eqn:representer_thm}
\widehat{F}=\sum_{i=1}^{n}k_{\mu}(\cdot,a_{i})\beta_{i}
\end{align}
for some functions $\{\beta_{i}\}_{i=1}^{n}\subset\cY$. In practice, finding
these $ n $ functions in the output space requires solving a block linear operator equation. For the high-dimensional PDE problems we consider in this work, solving such an equation may become prohibitively expensive from both operation count and memory required. A few workarounds were proposed in~\cite{kadri2016operator} such as certain diagonalizations, but these rely on simplifying assumptions that are somewhat limiting. More fundamentally, the representation of the solution in \cref{eqn:representer_thm} requires knowledge of the kernel $k_{\mu}$; in our setting we assume access only to the random feature pair $ (\varphi, \mu) $ which defines $k_{\mu}$ and not $k_{\mu}$ itself.

We thus explain how to make progress with this problem
given knowledge only of random features. Recall the empirical kernel given by \cref{eqn:kernel_empirical}, the RKHS $\cH_{k^{(m)}}$, and \cref{r:2}. The following result, proved in~\cite[Appendix A]{nelsen2021random}, shows that an RFM hypothesis class with a penalized least squares empirical loss function in optimization problem \cref{eqn:risk_empirical}--\cref{eqn:loss_square} is equivalent to kernel ridge regression \cref{eqn:general_leastsquares} restricted to $\cH_{k^{(m)}}$.

\begin{result}[random feature ridge regression is equivalent to a kernel method]\label{res:minimizer_equiv}
Assume that $\varphi \in L^2_{\nu \times \mu}(\cX \times \Theta; \cY)$
and that the random features $\{\varphi(\slot;\theta_j)\}_{j=1}^m$ are linearly independent in $L^2_{\nu}(\cX;\cY)$.
Fix $ \lambda\geq 0 $. Let $ \widehat{\al}\in\R^{m} $ be the unique minimum norm
solution of
\begin{align}\label{eqn:opt_randomfeature}
\min_{\al\in\R^{m}}\left\{\sum_{i=1}^{n}\dfrac{1}{2}\norm[\bigg]{y_i-\dfrac{1}{m}\sum_{l=1}^{m}\al_{l}\varphi(a_i;\theta_{l})}_{\cY}^{2}+\dfrac{\lambda}{2m}\norm*{\al}_{2}^{2}\right\}\, .
\end{align}
Then the RFM defined by this choice $\al=\widehat{\al}$ satisfies
\begin{align}\label{eqn:opt_equivalence}
F_{m}(\slot;\widehat{\al})=\argmin_{F\in\cH_{k^{(m)}}} \left\{\sum_{i=1}^{n}\dfrac{1}{2}\norm[\big]{y_i-F(a_i)}_{\cY}^{2}+\dfrac{\lambda}{2}\norm[\big]{F}_{\cH_{k^{(m)}}}^{2}\right\}\, .
\end{align}
\end{result}

Solving the convex problem~\cref{eqn:opt_randomfeature} trains the RFM. The first order condition for a global minimizer leads to the normal equations
\begin{align}\label{eqn:opt_normaleqn}
\sum_{j=1}^{m}\Biggl(\dfrac{1}{m}\sum_{i=1}^{n}\ip[\big]{\varphi(a_i;\theta_{l})}{\varphi(a_i;\theta_{j})}_{\cY} + \lambda \delta_{lj}\Biggr)\al_{j} =\sum_{i=1}^{n}\ip[\big]{\varphi(a_i;\theta_{l})}{y_i}_{\cY}
\end{align}
for each $l\in\{1,\ldots,m\}$, where $\delta_{lj}=1$ if $l=j$ and equals zero otherwise. This is an $ m $-by-$ m $ linear system of equations for $ \al\in\R^{m} $ that is standard to solve. In the case $ \lambda=0 $, the minimum norm solution of \cref{eqn:opt_normaleqn} may be written in terms of a pseudoinverse operator (see \cite[sect.~6.11]{luenberger1997optimization}).

\Cref{eqn:opt_normaleqn} reveals that the trained RFM $F_m(\slot;\widehat{\al})$ is a linear function of the labeled output data $\{y_i\}_{i=1}^n$. This property is undesirable from the perspective of statistical optimality. Indeed, it is known that any estimator that is linear in the output training data is minimax \emph{suboptimal} for certain classes of problems \cite[Theorem 1, sect. 4.1, p. 6]{suzuki2020benefit}. However, any adaptation of the feature pair $(\varphi,\mu)$ to the training data will break this property and potentially restore optimality. For example, choosing $\lambda$ or hyperparameters appearing in $(\varphi,\mu)$ based on a cross validation procedure would make the RF pair data-dependent as desired. This is typically done in practice~\cite{dunbar2024hyperparameter}.

\begin{example}[Brownian bridge]\label{ex:bb}
        We now provide a one-dimensional instantiation of the RFM to illustrate the methodology. Take the input space as $ \cX\defeq (0,1) $, output space $ \cY\defeq \R $, input space measure $ \nu \defeq \mathsf{Unif}(0,1) $ to be uniform, and random parameter space $ \Theta\defeq \R^{\infty} $. Denote the input by $ a=x\in\cX $. Then, consider the random feature map $ \varphi\colon (0,1)\times \R^{\infty}\to \R $ defined by the \emph{Brownian bridge}
        \begin{align}\label{eqn:bb}
        \varphi(x;\theta)\defeq \sum_{j=1}^{\infty}\theta^{(j)}(j\pi)^{-1}\sqrt{2}\sin(j\pi x)\, , \qw \theta^{(j)}\diid N(0,1)\, ,
        \end{align}
        where $ \theta\defeq\{ \theta^{(j)} \}_{j\in\N} $ and $ \mu\defeq N(0,1)\times N(0,1)\times\cdots$. For any realization of $ \theta \sim \mu$, the function $ \varphi(\slot;\theta) $ is a Brownian motion constrained to zero at $ x=0 $ and $ x=1 $. The induced kernel $ k_{\mu}\colon(0,1)\times (0,1)\to\R $ is then simply the covariance function of this stochastic process:
        \begin{align}\label{eqn:bb_kernel}
        k_{\mu}(x,x')=\E^{\theta\sim\mu}\bigl[\varphi(x;\theta)\varphi(x';\theta)\bigr]=\min\{x,x'\}-xx'\, .
        \end{align}
        Note that $k_{\mu}$ is the Green's function for the negative Laplacian on $(0,1)$ with Dirichlet boundary conditions. Using this fact, we may explicitly characterize the associated RKHS $ \cH_{k_{\mu}} $ as follows. First, we have
        \begin{align}\label{eqn:bb_covariance}
        T_{k_{\mu}}f = \int_{0}^{1}k_{\mu}(\cdot, y)f(y)\dd{y}=\biggl(-\frac{d^2}{dx^2}\biggr)^{-1}f\, ,
        \end{align}
        where the negative Laplacian has domain $H^2((0,1);\R)\cap H^1_0((0,1);\R) $. Viewing $ T_{k_{\mu}} $ as an operator from $ L^{2}((0,1);\R) $ into itself, from~\cref{eqn:ip_rkhs} we conclude, upon integration by parts, that for any elements $f$ and $g$ of $\cH_{k_{\mu}}$, it holds that
        \begin{align}\label{eqn:bb_rkhsspace}
        \ip{f}{g}_{\cH_{k_{\mu}}}=\ip{f}{T_{k_{\mu}}^{-1}g}_{L^2}=\ip[\bigg]{\frac{df}{dx}}{\frac{dg}{dx}}_{L^2}=\ip{f}{g}_{H_0^1}\, .
        \end{align}
        Note that the last identity does indeed define an inner product on $H^1_0.$ By this formal argument we identify the RKHS $ \cH_{k_{\mu}} $ as the Sobolev space $ H_0^1((0,1);\R) $. Furthermore, the Brownian bridge may be viewed as the Gaussian measure $N(0,T_{k_{\mu}})$.
        \begin{figure}[tb]
                \centering
                \begin{subfigure}[]{0.49\textwidth}
                        \centering
                        \includegraphics[width=\textwidth]{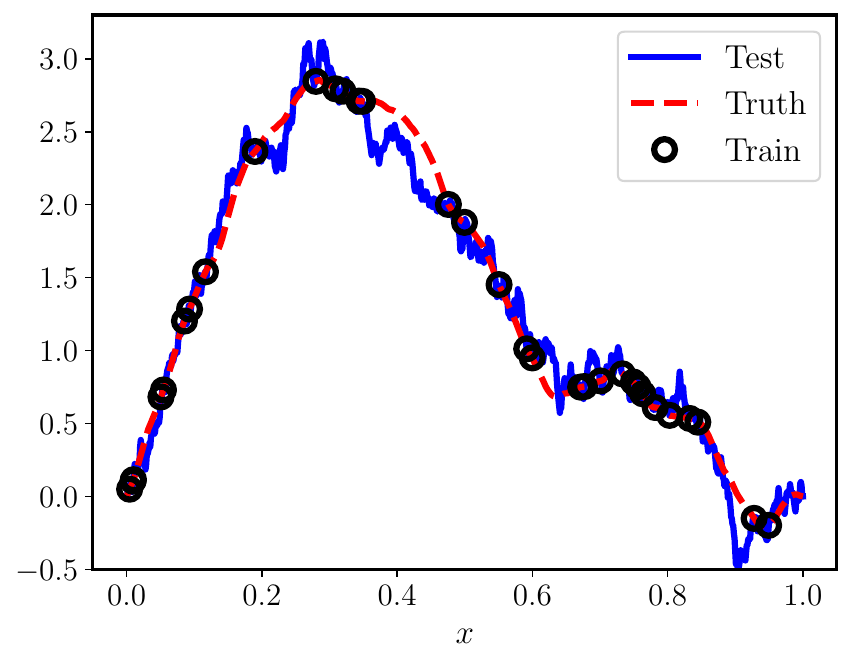}
                        \caption{$ m=50 $}
                        \label{fig:bb1}
                \end{subfigure}%
                \hfill%
                \begin{subfigure}[]{0.49\textwidth}
                        \centering
                        \includegraphics[width=\textwidth]{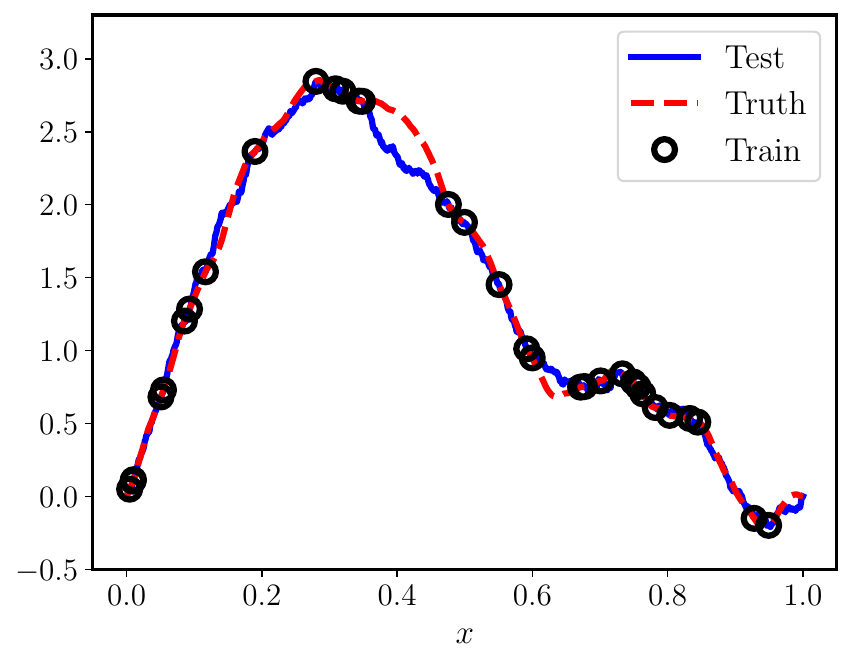}
                        \caption{$ m=500 $}
                        \label{fig:bb2}
                \end{subfigure}
                \begin{subfigure}[]{0.49\textwidth}
                        \centering
                        \includegraphics[width=\textwidth]{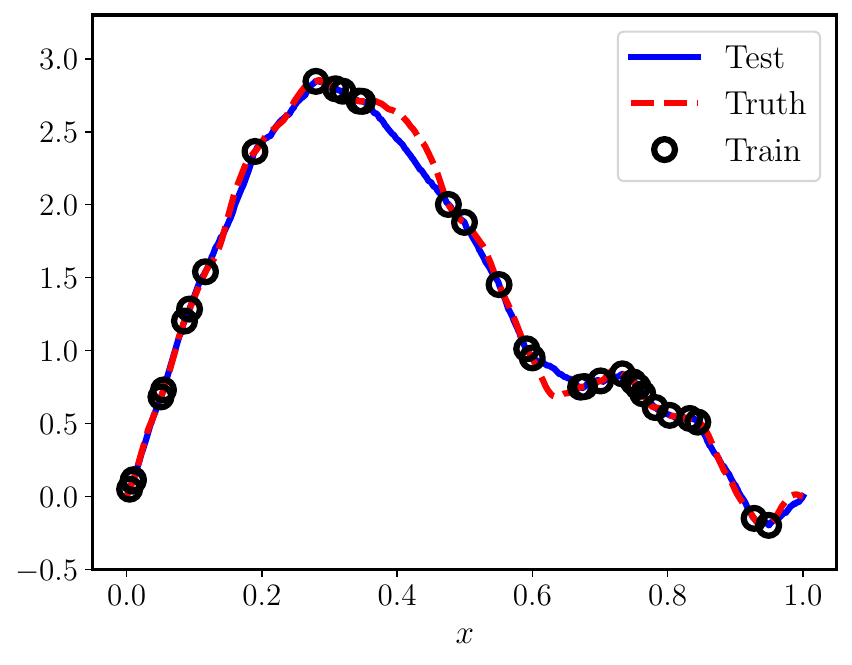}
                        \caption{$ m=5000 $}
                        \label{fig:bb3}
                \end{subfigure}%
                \hfill%
                \begin{subfigure}[]{0.49\textwidth}
                        \centering
                        \includegraphics[width=\textwidth]{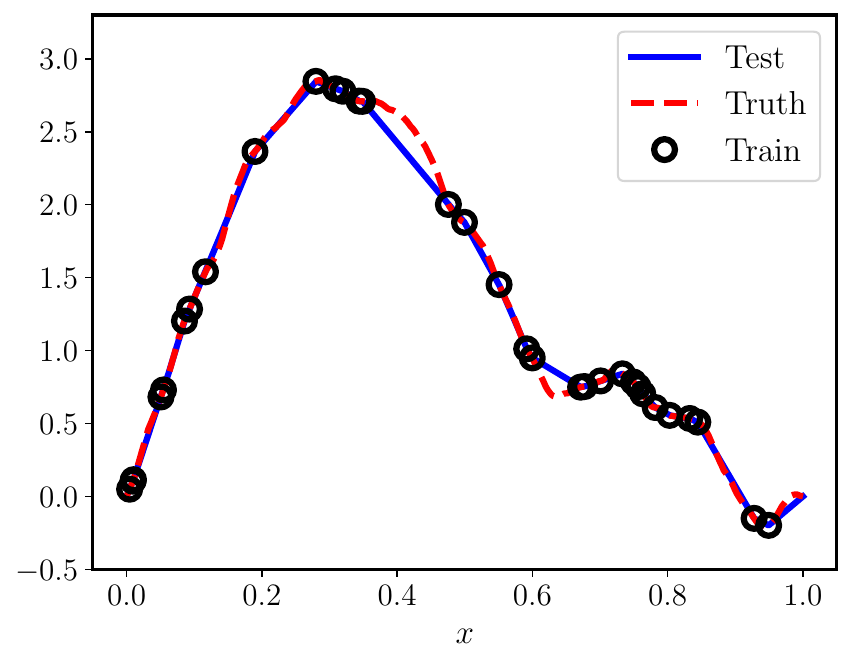}
                        \caption{$ m=\infty $}
                        \label{fig:bb4}
                \end{subfigure}
                \vspace{-5mm}
                \caption{Brownian bridge RFM for one-dimensional input-output spaces with $ n=32 $ training points fixed and $ \lambda=0 $~(\Cref{ex:bb}): As $ m\to\infty $, the RFM approaches the nonparametric interpolant given by the representer theorem~(Figure~\ref{fig:bb_compare}{\rm{(\subref{fig:bb4})}}), which in this case is a piecewise linear approximation of the true function (an element of RKHS $ \cH_{k_{\mu}}=H_{0}^1 $, shown in red). Blue lines denote the trained model evaluated on test data points and black circles denote evaluation at training points.}
                \label{fig:bb_compare}
        \end{figure}
        Approximation using the RFM with the Brownian bridge random features is illustrated in~\Cref{fig:bb_compare}. Since $ k_{\mu}(\cdot,x) $ is a piecewise linear function, a kernel interpolation or regression method will produce a piecewise linear approximation. Indeed, the figure indicates that the RFM with $ n $ training points fixed approaches the optimal piecewise linear kernel interpolant as $ m\to\infty $.
\end{example}

The Brownian bridge in \cref{ex:bb} illuminates a more fundamental idea. For this low-dimensional problem, an expansion in a deterministic Fourier sine basis would of course be more natural. But if we do not have a natural, computable orthonormal basis, then randomness provides a useful alternative representation; notice that the random features each include random combinations of the deterministic Fourier sine basis in this example. For the more complex problems that we study numerically in the next two sections, we lack knowledge of good, computable bases for general maps in infinite dimensions. The RFM approach exploits randomness to explore, implicitly discover the structure of, and represent such maps. Thus we now turn away from this example of real-valued maps defined on a subset of the real line and instead consider the use of random features to represent maps between spaces of functions. It turns out that theoretical guarantees are still possible to obtain in this setting.

\subsection{Error Bounds}\label{sec:theory}
In this subsection, we review a recent comprehensive error analysis~\cite{lanthaler2023error} of the random feature ridge regression problem~\cref{eqn:opt_equivalence} in the general infinite-dimensional input and output space setting. This is the sharpest available theory for misspecified problems. Owing to its tractable optimization, the RFM is one of the first guaranteed convergent operator learning algorithms for nonlinear problems that is actually implementable on a computer with controlled complexity.
To see this more concretely, we require the following technical assumptions.
\begin{assumption}[data and features]\label{ass:theory}
The following hold true.
    \begin{enumerate}[label=(\roman*)]
        \item The ground truth operator $\Fd$ satisfies $\Fd\in L^\infty_\nu(\cX;\cY)$.
        \item The noise-free training data are given by $a_i\diid\nu$ and $y_i=\Fd(a_i)$ for each $i$.
        \item The random feature map $\varphi\in L^\infty_{\nu\times \mu}(\cX\times\Theta;\cY)$ is measurable and bounded.
        \item The RKHS $\cH_{k_\mu}$ corresponding to the pair $(\varphi,\mu)$ is separable.
    \end{enumerate}
\end{assumption}

Our first convergence result is qualitative and follows from \cite[Theorem 3.10, p. 6]{lanthaler2023error}, which itself is a consequence of a more general error estimate \cite[Theorem 3.4, pp. 4--5]{lanthaler2023error}.\footnote{The regularization parameter $\lambda$ in \cref{thm:converge} and \cref{sec:opt} is equal to $n$ times the regularization parameter that is discussed in \cite{lanthaler2023error}, which is also denoted by the same symbol.}

\begin{theorem}[almost sure convergence of trained RFM]\label{thm:converge}
    Let \cref{ass:theory} hold.
    Suppose that the integral operator $T_{k_\mu}\in\cL(L^2_\nu(\cX;\cY))$ in \cref{eqn:integral_operator} is injective. Let $\{\delta_l\}_{l\in\N}\subset (0,1)$ be any positive sequence with the property that $\sum_{l=1}^\infty \delta_l <\infty$. For $l\in\N$, denote by $\widehat{\al}^{(l)}\in\R^{m_l}$ the trained RFM coefficients corresponding to \cref{eqn:opt_randomfeature} with $m=m_l$ random features, $n=n_l$ training samples, and regularization parameter $\lambda=\lambda_l$. If
    \begin{align}\label{eqn:complexity_seq}
        m_l\simeq \delta_l^{-1}\log(2/\delta_l)\,,\quad n_l\simeq \delta_l^{-2}\log(2/\delta_l)\,, \qa \lambda_l\simeq m_l\,,
    \end{align}
    then the trained RFM satisfies
    \begin{align}\label{eqn:error_seq}
        \P\biggl\{\lim_{l\to\infty}\E^{a\sim\nu}\norm[\big]{\Fd(a)-F_{m_l}(a;\widehat{\al}^{(l)})}_{\cY}^2 = 0 \biggr\} = 1\,.
    \end{align}
\end{theorem}

The probability in \cref{eqn:error_seq} is w.r.t. the joint law of the data $\{a_i\}_{i=1}^n\sim\nu^{\otimes n}$ and the feature parameters $\{\theta_j\}_{j=1}^M\sim \mu^{\otimes m}$. Going beyond the existence of an accurate approximation to $\Fd$, \cref{thm:converge} shows that the random feature ridge regression algorithm delivers a strongly consistent statistical estimator of $\Fd$ in the limit of large $m$, $n$, and $\lambda$. That is, the trained RFM that one actually obtains in practice converges (along a subsequence w.r.t. $n$) to the true underlying operator $\Fd$ with probability one. The three quantities $m$, $n$, and $\lambda$ are linked via a summable sequence $\{\delta_l\}$, which determines how they are simultaneously sent to infinity. The conditions of the theorem are satisfied with $\delta_l=l^{-2} \to 0$, for example.

The next theorem delivers a high probability nonasymptotic error bound that includes both parameter and sample complexity contributions that only depend algebraically on the reciprocal of the error instead of exponentially~\cite[cp. sect. 5]{kovachki2024operator}. It is a consequence of \cite[Theorem 3.7, p. 5]{lanthaler2023error} and controls sources of error due to regularization, finite parametrization, finite data, and optimization.
\begin{theorem}[complexity bounds for trained RFM]\label{thm:rate}
    Let $\ep\in (0,1)$ be an arbitrary error tolerance. Let $\widehat{\al}\in\R^m$ denote the trained RFM coefficients from \cref{eqn:opt_randomfeature} with training sample size $n\in\N$ and regularization parameter $\lambda\in (0,n)$. Suppose that $\Fd$ belongs to the RKHS $\cH_{k_\mu}$ corresponding to the random feature pair $(\varphi,\mu)$. Under \cref{ass:theory}, there exists an absolute constant $C>0$ such that if
    \begin{align}\label{eqn:complexity_bound}
        m\geq 11\ep^{-2}\,, \quad n\geq 10\ep^{-4}\,, \qa \lambda\leq 10\ep^{-2}\,,
    \end{align}
    then the trained RFM $F_m(\slot;\widehat{\al})$ satisfies the high probability $L^2_\nu(\cX;\cY)$ error bound
    \begin{align}\label{eqn:error_bound}
        \P\biggl\{\sqrt{\E^{a\sim\nu}\norm[\big]{\Fd(a)-F_m(a;\widehat{\al})}_{\cY}^2} \leq \bigl(C\norm{\Fd}_{\cH_{k_\mu}}\bigr)\ep \biggr\}\geq 0.999\,.
    \end{align}
\end{theorem}

The takeaway from \cref{thm:rate} is that, up to constant factors, an appropriately tuned regularization parameter $\lambda\simeq\sqrt{n}$ and number of random features $m\simeq \sqrt{n}$ are enough to guarantee a trained RFM generalization error of size $n^{-1/4}\simeq m^{-1/2}$ with high probability. However, this quantitative result is dependent on the well-specified condition $\Fd\in\cH_{k_\mu}$, which is quite difficult to verify in practice. It would be interesting to identify concrete operators of interest that actually belong to such RKHSs. Similar questions are also open for the Barron~\cite{e2022barron} and operator Barron spaces~\cite{korolev2021two} that correspond to NN models instead of RFMs.

The parameter complexity bound $m\gtrsim \ep^{-2}$ in \cref{eqn:complexity_bound} corresponds to the standard ``Monte Carlo'' parametric rate of estimation. Due to the i.i.d. sampling in~\cref{def:rfm} of the RFM, we expect this parametric rate to be sharp. However, the sample complexity bound $n\gtrsim \ep^{-4}$ from \cref{eqn:complexity_bound} is likely not sharp for fixed $\Fd$. Indeed, it is a worst case bound~\cite{caponnetto2007optimal} that presumably can be improved to $n\gtrsim \ep^{-(2+\delta)}$ for some small $\delta>0$ under stronger assumptions; see, e.g., \cite{rudi2017generalization} in the $\cY=\R$ setting. Such ``fast rates'' are empirically observable in numerical experiments.
We remark that the constants in \cref{thm:rate} were not optimized and could be improved. Additional refinements to \cref{thm:converge,thm:rate} that account for discretization error, noisy output data, and smoothness misspecification may be found in \cite[sect. 3]{lanthaler2023error}.

\section{Application to PDE Solution Operators}
\label{sec:application}
In this section, we design the random feature maps $ \varphi\colon \cX\times\Theta\to\cY $ and measures $ \mu $ for the RFM approximation of two particular PDE parameter-to-solution maps: the evolution semigroup of the viscous Burgers' equation in \cref{sec:burg_formulation} and the coefficient-to-solution operator for the Darcy problem in \cref{sec:darcy_formulation}. It is well known to kernel method practitioners that the choice of kernel (which in this work follows from the choice of $ (\varphi,\mu) $) plays a central role in the quality of the function reconstruction. While our method is purely data-driven and requires no knowledge of the governing PDE, we take the view that any prior knowledge can, and should, be introduced into the design of $ (\varphi, \mu)$. However, the question of how to automatically determine good random feature pairs for a particular problem or dataset, inducing data-adapted kernels, is open. The maps $ \varphi$ that we choose to employ are nonlinear in both arguments. We also detail the probability measure $ \nu $ on the input space $ \cX $ for each of the two PDE applications; this choice is crucial because while we desire the trained RFM to transfer to arbitrary out-of-distribution inputs from $ \cX $, we can in general only expect the learned map to perform well when restricted to inputs statistically similar to those sampled from $ \nu $.

\subsection{Burgers' Equation: Formulation}\label{sec:burg_formulation}
The viscous Burgers' equation in one spatial dimension is representative of the advection-dominated PDE problem class in some regimes; these time-dependent equations are not conservation laws due to the presence of small dissipative terms, but nonlinear transport still plays a central role in the evolution of solutions. The initial value problem we consider is
\begin{align}\label{eqn:burgers_ibvp}
\begin{cases}
\begin{alignedat}{2}
\frac{\partial u}{\partial t}+\frac{\partial}{\partial x}\left(\frac{u^2}{2}\right)-\ep \frac{\partial^2 u}{\partial x^2}&=f && \qin (0,\infty)\times(0,1)\,,\\
u(\cdot, 0)=u(\cdot, 1)\, ,\quad \frac{\partial u}{\partial x}(\cdot, 0)&=\frac{\partial u}{\partial x}(\cdot, 1) && \qin (0,\infty)\,,\\
u(0,\cdot)&=a && \qin (0,1)\, ,
\end{alignedat}
\end{cases}
\end{align}
where $ \ep>0 $ is the viscosity (i.e., diffusion coefficient) and we have imposed periodic boundary conditions. The initial condition $ a $ serves as the input and is drawn according to a Gaussian measure defined by
\begin{align}\label{eqn:prior_burgers}
a\sim \nu\defeq N(0,\cC)
\end{align}
with Mat\'ern-like covariance operator \cite{dunlop2017hierarchical,matern2013spatial}
\begin{align}\label{eqn:prior_covariance}
\cC\defeq\tau^{2\al-d}(-\lap +\tau^{2}\id)^{-\al}\, ,
\end{align}
where $d=1$ and the negative Laplacian $ -\lap $ is defined over the torus $\mathbb{T}^1=[0,1]_{\mathrm{per}}$ and restricted to functions which integrate to zero over $ \mathbb{T}^1 $. The hyperparameter $ \tau\geq 0 $ is an inverse length scale  and $ \al>1/2 $ controls the regularity of the draw. Such $ a $ are almost surely H\"older and Sobolev regular with exponent up to $ \al - 1/2 $ \cite[Theorem 12, p. 338]{Dashti2017}, so in particular $ a\in \cX\defeq L^{2}(\mathbb{T}^1;\R) $. Then for all $ \ep>0 $, the unique global solution $ u(t,\cdot) $ to~\cref{eqn:burgers_ibvp} is real analytic for all $ t>0 $~\cite[Theorem 1.1]{kiselev2008blow}. Hence, setting the output space to be $ \cY\defeq  H^{s}(\mathbb{T}^1;\R) $ for any $ s>0 $, we may define the solution map
\begin{align}\label{eqn:solnmap_burg}
\begin{split}
\Fd\colon L^2 &\to H^{s}\, , \\
a&\mapsto \Fd(a)\defeq \Psi_{T}(a)=u(T,\cdot)\, ,
\end{split}
\end{align}
where $ \{\Psi_{t}\}_{t>0} $ forms the solution operator semigroup for~\cref{eqn:burgers_ibvp} and we fix the final time $ t=T>0 $. The map $ \Fd $ is smoothing and nonlinear.

\begin{figure}[tb]
        \centering
        \begin{subfigure}[]{0.49\textwidth}
                \centering
                \includegraphics[width=\textwidth]{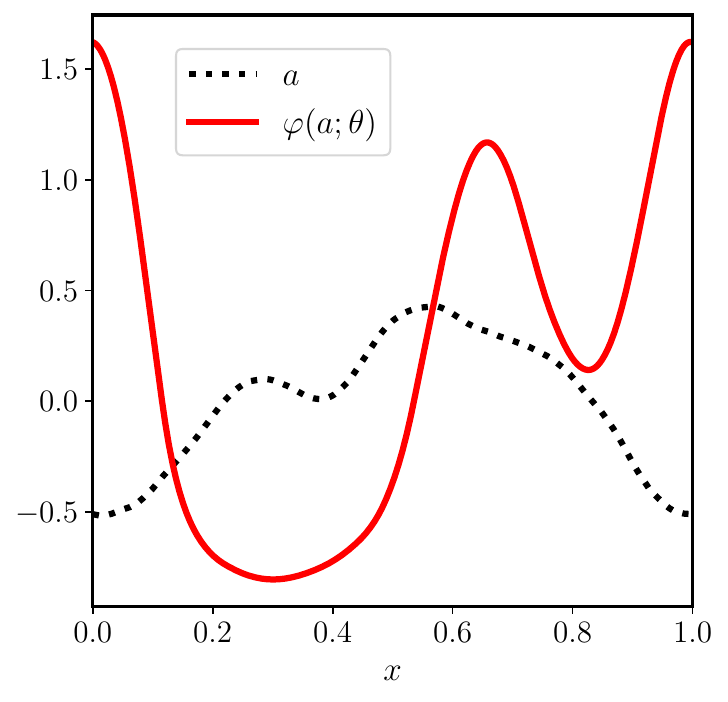}
                \caption{Random feature}
                \label{fig:rf_sample_burg}
        \end{subfigure}%
        \hfill%
        \begin{subfigure}[]{0.49\textwidth}
                \centering
                \includegraphics[width=\textwidth]{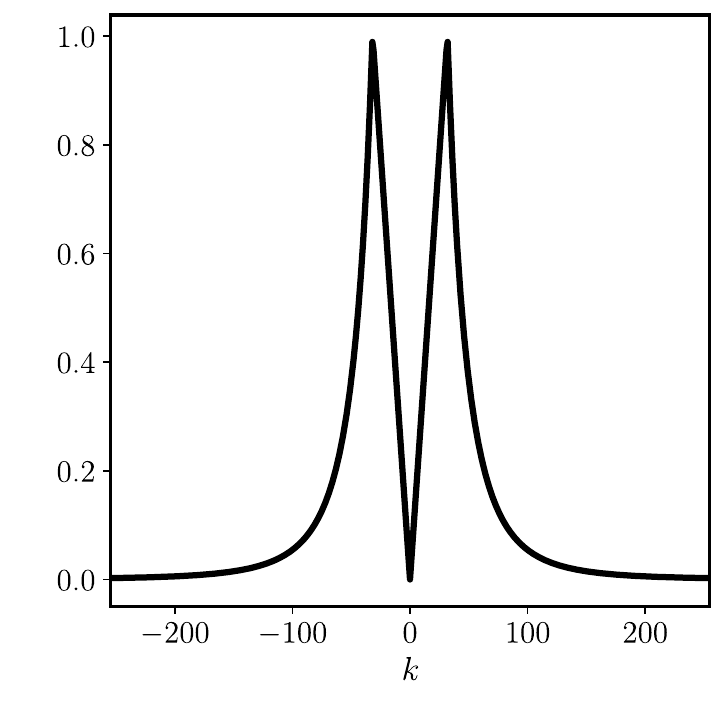}
                \caption{Filter function}
                \label{fig:filter_func1}
        \end{subfigure}
        \vspace{-5mm}
        \caption{Random feature map construction for Burgers' equation: Figure~\ref{fig:rf_and_filter}{\rm{(\subref{fig:rf_sample_burg})}} displays a representative input-output pair for the random feature $ \varphi(\slot;\theta) $ with $ \theta\sim\mu $~\cref{eqn:rf_fourier}, while Figure~\ref{fig:rf_and_filter}{\rm{(\subref{fig:filter_func1})}} shows the filter $k\mapsto \chi(k) $ for $ \delta=0.0025 $ and $ \beta=4 $~\cref{eqn:filter}.}
        \label{fig:rf_and_filter}
\end{figure}

We now describe a random feature map for use in the RFM~\cref{eqn:rfm} that we call \emph{Fourier space random features}. Let $ \cF $ denote the Fourier transform over spatial domain $ \mathbb{T}^1 $ and define $ \varphi\colon \cX\times\Theta\to\cY $ by
\begin{align}\label{eqn:rf_fourier}
\varphi(a;\theta)\defeq\sigma\left(\cF^{-1}(\chi\cF a\cF \theta)\right)\, ,
\end{align}
where $\sigma(\slot)$, the $ \ON{ELU} $ function defined below, is defined as a mapping on $\mathbb{R}$ and applied pointwise to functions. Viewing $ \Theta\subseteq L^{2}(\mathbb{T}^1;\R) $, the randomness enters through $ \theta\sim \mu\defeq N(0,\cC') $ with $ \cC' $ the same covariance operator as in~\cref{eqn:prior_covariance} but with potentially different inverse length scale and regularity, and the \emph{wavenumber filter function} $ \chi\colon\Z\to\R_{\geq 0} $ is given for $k\in\Z$ by
\begin{align}\label{eqn:filter}
\chi(k)\defeq \sigma_{\chi}(2\pi\abs{k}\delta)\, , \qw \sigma_{\chi}(r)\defeq \max\Bigl(0, \min\bigl(2r, (r+1/2)^{-\beta}\bigr)\Bigr)\,,
\end{align}
$ \delta>0$,  and $\beta>0 $. The map $ \varphi(\slot;\theta) $ essentially performs a filtered random convolution with the initial condition. Figure~\ref{fig:rf_and_filter}{\rm{(\subref{fig:rf_sample_burg})}}  illustrates a sample input and output from $ \varphi $. Although simply hand-tuned for performance and not optimized, the filter $ \chi $ is designed to shuffle energy in low to medium wavenumbers and cut off high wavenumbers~(see~Figure~\ref{fig:rf_and_filter}{\rm{(\subref{fig:filter_func1})}}), reflecting our prior knowledge of solutions to~\cref{eqn:burgers_ibvp}.

We choose the activation function $ \sigma $ in~\cref{eqn:rf_fourier} to be the exponential linear unit
\begin{align}\label{eqn:activation_elu}
r\mapsto \ON{ELU}(r)\defeq
\begin{cases}
\begin{alignedat}{2}
r&\, , \ \ \ && \text{if }\,r\geq 0\, ,\\
e^{r}-1&\, , \ \ \ && \text{if }\,r<0\, .
\end{alignedat}
\end{cases}
\end{align}
The $ \ON{ELU} $ function has successfully been used as activation in other  machine learning frameworks for related nonlinear PDE problems~\cite{lee2020model, patel2018nonlinear,patel2020physics}. We also find $ \ON{ELU}(\slot) $ to perform better in the RFM framework over several other choices including $ \ON{ReLU}(\slot)$, $\tanh(\slot)$, $\ON{sigmoid}(\slot)$, $\sin(\slot) $, $ \ON{SELU}(\slot) $, and $ \ON{softplus}(\slot) $. Note that the pointwise evaluation of the $ \ON{ELU} $ function in~\cref{eqn:rf_fourier} will be well defined, by Sobolev embedding, for $ s>1/2 $ sufficiently large in the definition of $ \cY=H^{s} $. Since the solution operator maps into $H^s$ for any $s>0$, this does not constrain the method.

\subsection{Darcy Flow: Formulation}\label{sec:darcy_formulation}
\begin{figure}[tb]
        \centering
        \begin{subfigure}[]{0.49\textwidth}
                \centering
                \includegraphics[width=\textwidth]{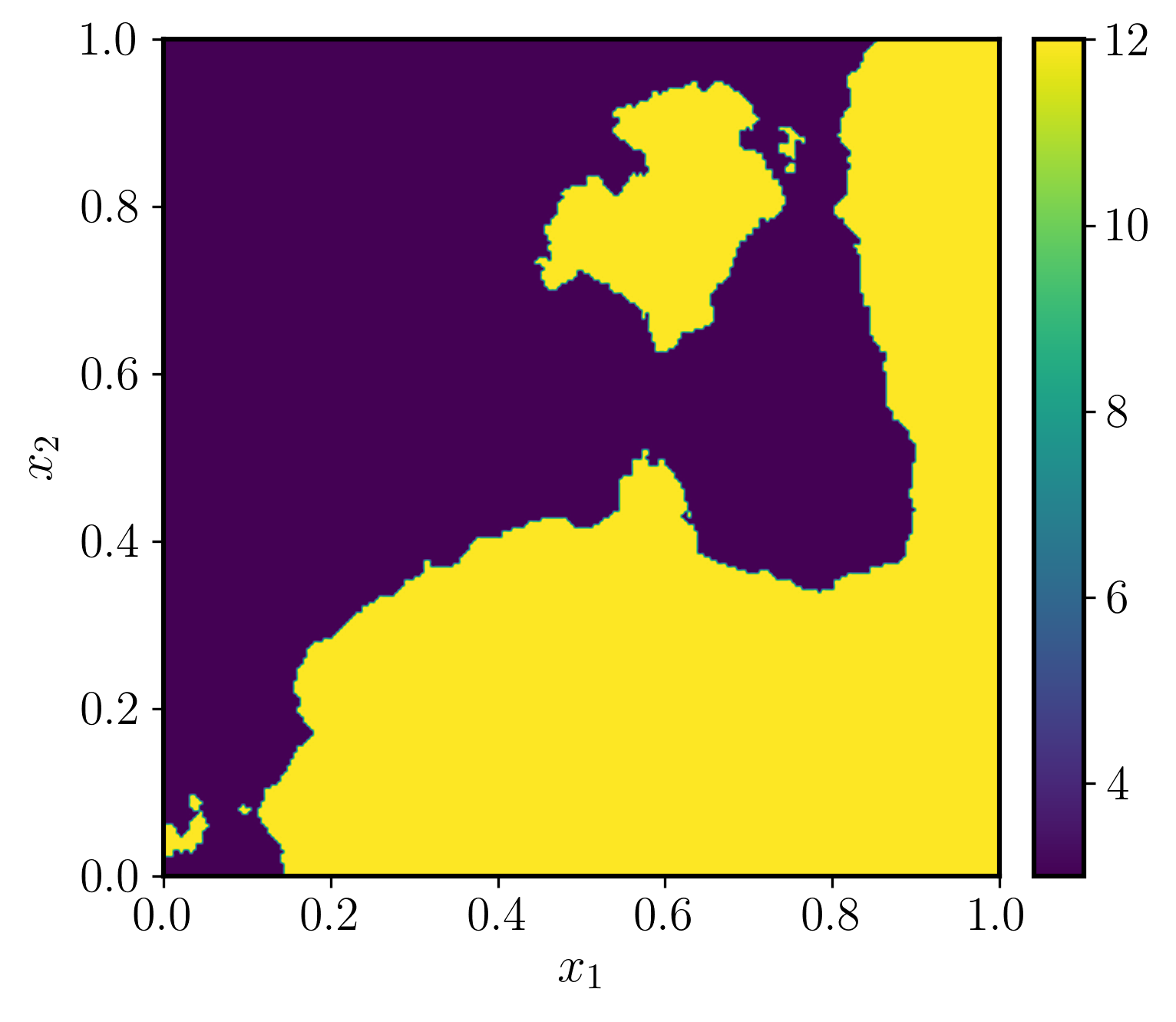}
                \caption{$ a\sim \nu $}
                \label{fig:coef_darcy}
        \end{subfigure}%
        \hfill%
        \begin{subfigure}[]{0.49\textwidth}
                \centering
                \includegraphics[width=\textwidth]{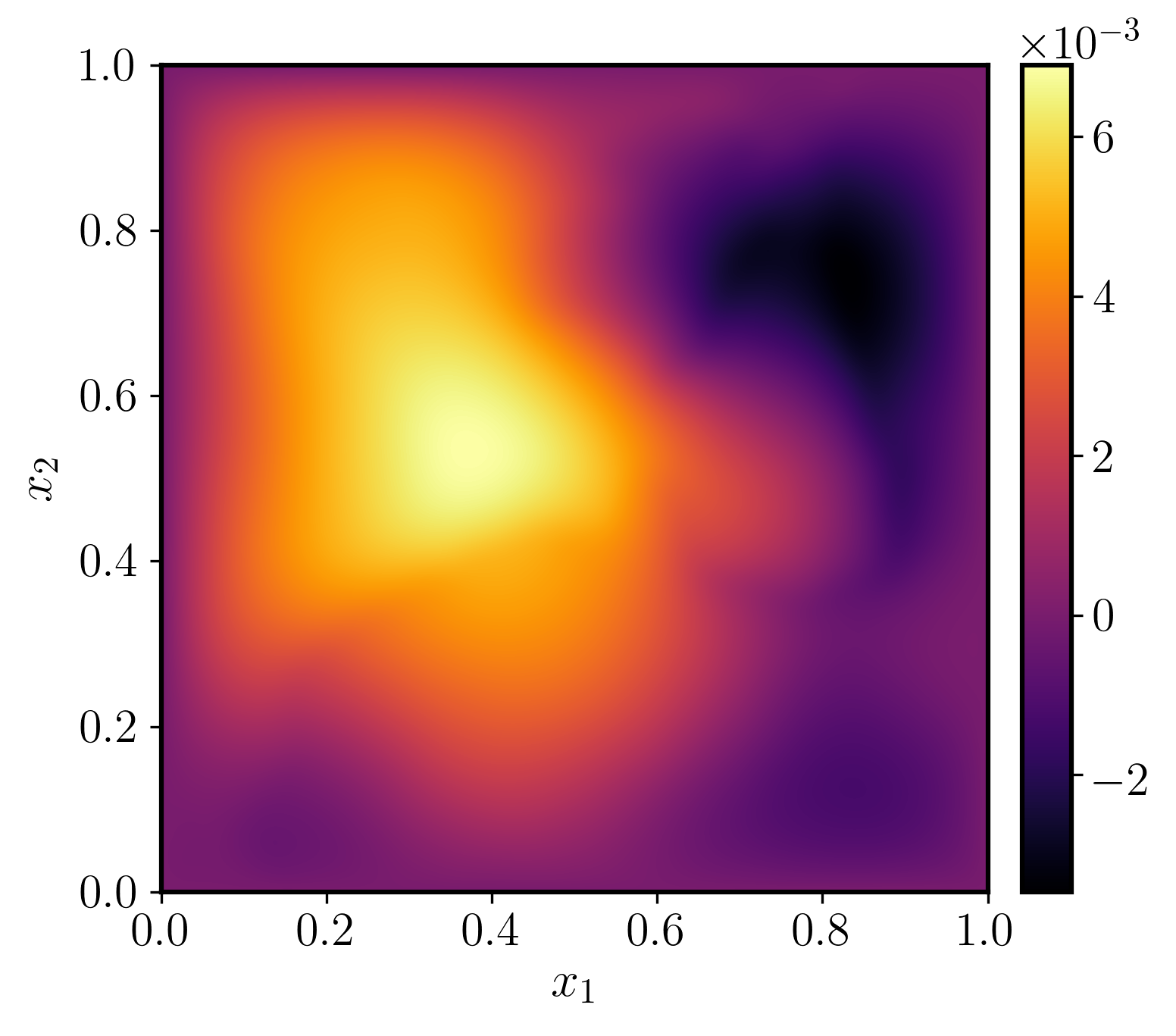}
                \caption{$ \varphi(a;\theta), \qw \theta\sim\mu$}
                \label{fig:rf_darcy}
        \end{subfigure}
        \vspace{-5mm}
        \caption{Random feature map construction for Darcy flow: Figure~\ref{fig:rf_and_coef_darcy}{\rm{(\subref{fig:coef_darcy})}} displays a representative input draw $ a $ with $ \tau=3,\, \al=2 $  and $ a^{+}=12,\, a^{-}=3 $; Figure~\ref{fig:rf_and_coef_darcy}{\rm{(\subref{fig:rf_darcy})}} shows the output random feature $ \varphi(a;\theta) $~(equation~\ref{eqn:rf_predictor_corrector}) taking the coefficient $ a $ as input. Here, $ f\equiv 1 $, $ \tau'=7.5,\, \al'=2 $, $ s^{+}=1/a^{+} $, $ s^{-}=-1/a^{-} $, and $ \delta = 0.15 $.}
        \label{fig:rf_and_coef_darcy}
\end{figure}

Divergence form elliptic equations~\cite{gilbarg2015elliptic} arise in a variety of applications, in particular, the groundwater flow in a porous medium governed by Darcy's law~\cite{bear2012fundamentals}. This linear elliptic boundary value problem reads
\begin{align}\label{eqn:darcy_flow}
\begin{cases}
\begin{alignedat}{2}
-\nabla\cdot(a\nabla u)&=f  && \qin D\,,\\
u&=0 && \qon \partial D\, ,
\end{alignedat}
\end{cases}
\end{align}
where $ D $ is a bounded open subset in $ \R^{d} $, $f$ represents sources
and sinks of fluid, $a$ the permeability of the porous medium, and $u$ the piezometric head; all three functions map $D$ into $\R$ and, in addition, $a$ is strictly positive almost everywhere in $D$. We work in a setting where $f$ is fixed and consider the input-output map defined by $a \mapsto u$. The measure $\nu$ on $a$ is a high contrast level set prior
constructed as the pushforward of a Gaussian measure:
\begin{align}\label{eqn:prior_levelset}
a\sim \nu\defeq \psi_{\sharp}N(0,\cC)\, .
\end{align}
Here $ \psi \colon \R\to\R $ is a threshold function defined for $r\in\R$ by
\begin{align}\label{eqn:prior_function_push}
\psi(r)\defeq a^{+}\one_{(0,\infty)}(r)+a^{-}\one_{(-\infty,0)}(r)\, ,\qw 0<a^{-}\leq a^{+}<\infty\, ,
\end{align}
applied pointwise to functions, and the covariance operator $ \cC $ is given in~\cref{eqn:prior_covariance} with $ d=2 $ and homogeneous Neumann boundary conditions on $ -\lap $. That is, the resulting coefficient $ a $ almost surely takes only two values ($ a^{+} $ or $ a^{-} $) and, as the zero level set of a Gaussian random field, exhibits random geometry in the physical domain $ D $. It follows that $ a\in L^{\infty}(D;\R_{\geq 0}) $ almost surely. Further, the size of the contrast ratio $ a^{+}/a^{-} $ measures the scale separation of this elliptic problem and hence controls the difficulty of reconstruction~\cite{bernardi2000adaptive}. See~Figure~\ref{fig:rf_and_coef_darcy}{\rm{(\subref{fig:coef_darcy})}} for a representative sample.

Given $ f\in L^{2}(D;\R) $, the standard Lax--Milgram theory may be applied to show that for coefficient $a\in\cX\defeq L^{\infty}(D;\R_{\geq 0})$, there exists a unique weak solution $ u\in \cY\defeq H_{0}^{1}(D;\R) $ for~\cref{eqn:darcy_flow}~(see, e.g.,~Evans~\cite{evans2010partial}). Thus, we define the ground truth solution map
\begin{align}\label{eqn:solnmap_darcy}
\begin{split}
\Fd\colon L^{\infty} &\to H_{0}^{1}\, , \\
a&\mapsto \Fd(a)\defeq u\, .
\end{split}
\end{align}
Although the PDE~\cref{eqn:darcy_flow} is linear, the solution map $ \Fd $ is nonlinear.

We now describe the chosen random feature map for this problem, which we call \emph{predictor-corrector random features}. Define $ \varphi\colon \cX\times\Theta\to\cY$ by $ \varphi(a;\theta)\defeq p_1 $ such that
\begin{subequations}\label{eqn:rf_predictor_corrector}
        \begin{align}
        -\lap p_0&=\dfrac{f}{a}+\sigma_{\gamma}(\theta_1)\,,\label{eqn:predictor}\\
        -\lap p_1&=\dfrac{f}{a}+\sigma_{\gamma}(\theta_2)+\nabla(\log a)\cdot\nabla p_0\, ,\label{eqn:corrector}
        \end{align}
\end{subequations}
where the boundary conditions are homogeneous Dirichlet, $ \theta=(\theta_1,\theta_2) \sim \mu\defeq\mu'\times\mu'$ are two Gaussian random fields each drawn from $ \mu'\defeq N(0, \cC') $, $ f $ is the source term in~\cref{eqn:darcy_flow}, and $ \gamma =(s^{+}, s^{-}, \delta)$ are parameters for a thresholded sigmoid $ \sigma_{\gamma}\colon\R\to\R $,
\begin{align}\label{eqn:sigmoidal_func_defn}
r\mapsto \sigma_{\gamma}(r)\defeq \dfrac{s^{+}-s^{-}}{1+e^{-r/\delta}}+s^{-}\, ,
\end{align}
and extended as a Nemytskii operator when applied to $\theta_1(\slot)$
or $\theta_2(\slot)$. We view $ \Theta\subseteq L^2(D;\R)\times L^2(D;\R) $. In practice, since $ \nabla a $ is not well defined when drawn from the level set measure, we replace $ a $ with $a_{\ep} $, where $ a_{\ep} \defeq v(1, \cdot) $ is a smoothed version of $ a $ obtained by evolving the following linear heat equation for one time unit:
\begin{align}\label{eqn:heat_mollify_neumann}
\begin{cases}
\begin{alignedat}{2}
\frac{\partial v}{\partial t}&=\eta\lap v && \qin (0,1)\times D\,,\\[6pt]
\mathsf{n}\cdot \nabla v &= 0 &&\qon(0,1)\times\partial D\,,\\[3pt]
v(0,\cdot) &= a &&\qin D\,,
\end{alignedat}
\end{cases}
\end{align}
where $ \mathsf{n} $ is the outward unit normal vector to $ \partial D $. An example of the response $ \varphi(a;\theta) $ to a piecewise constant input $ a\sim\nu$ is shown in~\Cref{fig:rf_and_coef_darcy} for some $ \theta\sim\mu $.

We remark that by removing the two random terms involving $ \theta_1$ and $ \theta_2 $ in~\cref{eqn:rf_predictor_corrector}, we obtain a remarkably accurate surrogate model for the PDE. This observation is representative of a more general iterative method, a predictor-corrector type iteration, for solving the Darcy equation~\cref{eqn:darcy_flow}, whose convergence depends on the size of $ a $. The map $ \varphi $ is essentially a random perturbation of a single step of this iterative method: \cref{eqn:predictor} makes a coarse prediction of the output, then~\cref{eqn:corrector} improves this prediction with a correction term derived from expanding the original PDE. This choice of $ \varphi $ falls within an ensemble viewpoint that the RFM may be used to improve preexisting surrogate models by taking $ \varphi(\slot;\theta) $ to be an existing emulator, but randomized in a principled way through $ \theta\sim\mu $.

For this particular example, we are cognizant of the facts that the random feature map $ \varphi $ requires full knowledge of the Darcy equation and a na\"ive evaluation of $ \varphi $ may be as expensive as solving the original PDE, which is itself a linear PDE; however, we believe that the ideas underlying the random features used here are intuitive and suggestive of what is possible in other applications areas. For example, RFMs may be applied on larger domains with simple geometries, viewed as supersets of the physical domain of interest, enabling the use of efficient algorithms such as the fast Fourier transform (FFT) even though these may not be available on the original problem, either because the operator to be inverted is spatially inhomogeneous or because of the complicated geometry of the physical domain.

\section{Numerical Experiments}
\label{sec:experiment}
We now assess the performance of our proposed methodology on the approximation of operators $ \Fd\colon\cX\to\cY $ presented in~\cref{sec:application}.
Practical implementation of the approach on a computer necessitates discretization of the input-output function spaces $ \cX $ and $ \cY $. Hence in the numerical experiments that follow, all infinite-dimensional objects such as the training data, evaluations of random feature maps, and random fields are discretized on an equispaced mesh with $ K $ grid points to take advantage of the $ O(K\log K) $ computational speed of the FFT. The simple choice of equispaced points does not limit the proposed approach, as our formulation of the RFM on function space allows the method to be implemented numerically with any choice of spatial discretization. Such a numerical discretization procedure leads to the problem of high- but finite-dimensional approximation of discretized target operators mapping $ \R^{K} $ to $ \R^{K} $ by similarly discretized RFMs. However, we emphasize the fact that $ K $ is allowed to vary, and we study the properties of the discretized RFM as $ K $ varies, noting that since the RFM is defined conceptually on function space in \cref{sec:problem} without reference to discretization, its discretized numerical realization has approximation quality consistent with the infinite-dimensional limit $ K\to\infty $. This implies that the same trained model can be deployed across the entire hierarchy of finite-dimensional spaces $ \R^{K} $ parametrized by $ K\in\N $ without the need to be retrained, provided
$K$ is sufficiently large. Thus in this section, our notation does not make explicit the dependence of the discretized RFM or target operators on mesh size $ K $. We demonstrate these claimed properties numerically.

The input functions and our chosen random feature maps~\cref{eqn:rf_fourier} and~\cref{eqn:rf_predictor_corrector} require i.i.d. draws of Gaussian random fields to be fully defined. We efficiently sample these fields by truncating a Karhunen--Lo\'eve expansion and employing fast summation of the eigenfunctions with FFT. More precisely, on a mesh of size $ K $, denote by $ g(\cdot) $ a numerical approximation of a Gaussian random field on domain $ D=(0,1)^{d} $, $ d=1,\, 2 $:
\begin{align}\label{eqn:grf_klexpansion}
g=\sum_{k\in Z_K}\xi_k\sqrt{\lambda_k}\phi_k \approx \sum_{k'\in\Z_{\geq 0}^{d}}\xi_{k'}\sqrt{\lambda_{k'}}\phi_{k'}  \sim N(0, \cC)\, ,
\end{align}
where $ \xi_{j} \sim N(0,1) $ i.i.d. for each $j$ and $ Z_K\subset\Z_{\geq 0} $ is a truncated one-dimensional lattice of cardinality $ K $ ordered such that $ \{\lambda_j\} $ is nonincreasing. The pairs $ (\lambda_{k'}, \phi_{k'}) $ are found by solving the eigenvalue problem $ \cC\phi_{k'}=\lambda_{k'}\phi_{k'} $ for nonnegative, symmetric, trace-class operator $ \cC $ \cref{eqn:prior_covariance}. Concretely, these solutions are given by
\begin{align}\label{eqn:eig_neumann}
\phi_{k'}(x)
=
\begin{cases}
\sqrt{2}\cos(k_1'\pi x_1)\cos(k_2'\pi x_2),& k_1' \ \text{or} \ k_2'=0\, ,\\
2\cos(k_1'\pi x_1)\cos(k_2'\pi x_2),& \text{otherwise}\, ,
\end{cases}\, \quad
\lambda_{k'}=\tau^{2\al - 2}(\pi^{2}\abs{k'}^{2}+\tau^{2})^{-\al},
\end{align}
for homogeneous Neumann boundary conditions when $ d=2 $, $ k'=(k_1', k_2')\in \Z_{\geq 0}^{2}{\setminus}\{0\} $, $ x=(x_1,x_2)\in (0,1)^{2} $, and given by
\begin{subequations}\label{eqn:periodic}
        \begin{align}
        \phi_{2j}(x)&=\sqrt{2}\cos(2\pi j x)\, ,\quad  \phi_{2j-1}(x)=\sqrt{2}\sin(2\pi j x)\, ,\quad \phi_0(x)=1\,,\\
        \lambda_{2j}&=\lambda_{2j-1}=\tau^{2\al - 1}(4\pi^{2}j^{2}+\tau^{2})^{-\al}\, ,\quad \lambda_0=\tau^{-1},
        \end{align}
\end{subequations}
for periodic boundary conditions when $ d=1 $, $ j\in\Z_{>0} $, and $ x\in (0,1) $. In both cases, we enforce that $ g $ integrate to zero over $ D $ by manually setting to zero the Fourier coefficient corresponding to multi-index $ k'=0 $. We use such $ g $ in all experiments that follow. Additionally, the $ k $ and $ k' $ used in this section to denote wavenumber indices should not be confused with our previous notation for kernels.

With the discretization and data generation setup now well defined, and the pairs $ (\varphi, \mu) $ given in \cref{sec:application}, the last algorithmic step is to train the RFM by solving~\cref{eqn:opt_normaleqn} and then test its performance. For a fixed number of random features $ m $, we only train and test a single realization of the RFM, viewed as a random variable itself. In each instance $ m $ is varied in the experiments that follow, the draws $ \{\theta_j\}_{j=1}^m $ are resampled i.i.d.~from $ \mu $.
To measure the distance between the trained RFM $ F_{m}(\cdot;\widehat{\al}) $ and the ground truth $ \Fd $, we employ the \emph{approximate expected relative test error}
\begin{align}\label{eqn:test_error}
e_{n',m}\defeq \dfrac{1}{n'}\sum_{j=1}^{n'}\dfrac{\norm{\Fd(a_j')-F_{m}(a_j';\widehat{\al})}_{L^2}}{\norm{\Fd(a_j')}_{L^2}} \approx \E^{a'\sim\nu}\left[ \dfrac{\norm{\Fd(a')-F_{m}(a';\widehat{\al})}_{L^2}}{\norm{\Fd(a')}_{L^2}}\right]\, ,
\end{align}
where the $\{a_j'\}_{j=1}^{n'}$ are drawn i.i.d.~from $\nu$ and $ n' $ denotes the number of input-output pairs used for testing. All $ L^2(D;\R) $ norms on the physical domain are numerically approximated by composite trapezoid rule quadrature. Since $ \cY\subset L^{2} $ for both the PDE solution operators \cref{eqn:solnmap_burg} and \cref{eqn:solnmap_darcy}, we also perform all required inner products during training in $ L^{2} $ rather than in $ \cY $; this results in smaller relative test error $ e_{n',m} $.

\subsection{Burgers' Equation: Experiment}\label{sec:burg_exp}
We generate a high resolution dataset of input-output pairs by solving Burgers' equation~\cref{eqn:burgers_ibvp} on an equispaced periodic mesh of size $ K=1025 $ (identifying the first mesh point with the last) with random initial conditions sampled from $ \nu=N(0,\cC) $ using \cref{eqn:grf_klexpansion}, where $ \cC $ is given
by~\cref{eqn:prior_covariance} with parameter choices $\tau=7$ and $\al=2.5$. The full order solver is an FFT-based pseudospectral method for spatial discretization \cite{fornberg1998practical} and a fourth order Runge--Kutta integrating factor time-stepping scheme for time discretization \cite{kassam2005fourth}. All data represented on mesh sizes $ K<1025 $ used in both training and testing phases are subsampled from this original dataset, and hence we consider numerical realizations of $ \Fd $ \cref{eqn:solnmap_burg} up to $ \R^{1025}\to\R^{1025} $. We fix $ n=512 $ training and $ n'=4000 $ testing pairs unless otherwise noted and also fix the viscosity to $ \ep=10^{-2} $ in all experiments. Lowering $ \ep $ leads to smaller length scale solutions and more difficult reconstruction; more data (higher $ n $) and features (higher $ m $) or a more expressive choice of $ (\varphi,\mu) $ would be required to achieve comparable error levels due to the slow decaying Kolmogorov width of the solution map. For simplicity, we set the forcing $ f\equiv 0 $, although nonzero forcing could lead to other interesting solution maps such as $ f\mapsto u(T,\cdot) $. It is easy to check that the solution will have zero mean for all time and a steady state of zero. Hence, we choose $ T\leq 2 $ to ensure that the solution is far enough away from steady state. For the random feature map~\cref{eqn:rf_fourier}, we fix the hyperparameters $ \al'=2 $, $ \tau'=5 $, $ \delta=0.0025 $, and $ \beta=4 $. The map itself is evaluated efficiently with the FFT and requires no other tools to be discretized. RFM hyperparameters were hand-tuned but not optimized. We find that regularization during training had a negligible effect for this problem, so the RFM is trained with $ \lambda=0 $ by solving the normal equations~\cref{eqn:opt_normaleqn} with the pseudoinverse to deliver the minimum norm least squares solution; we use the truncated SVD implementation in Python's $ \texttt{scipy.linalg.pinv2} $ for this purpose.

\begin{figure}[tb]
        \centering
        \begin{subfigure}[]{0.47\textwidth}
                \centering
                \includegraphics[width=\textwidth]{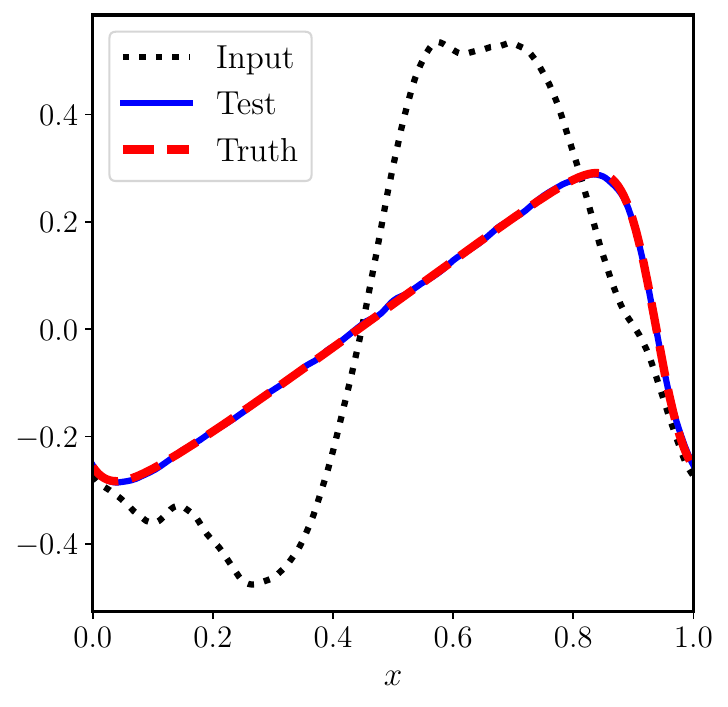}
                \caption{Input and output}
                \label{fig:prediction_onesample}
        \end{subfigure}%
        \hfill%
        \begin{subfigure}[]{0.49\textwidth}
                \centering
                \includegraphics[width=\textwidth]{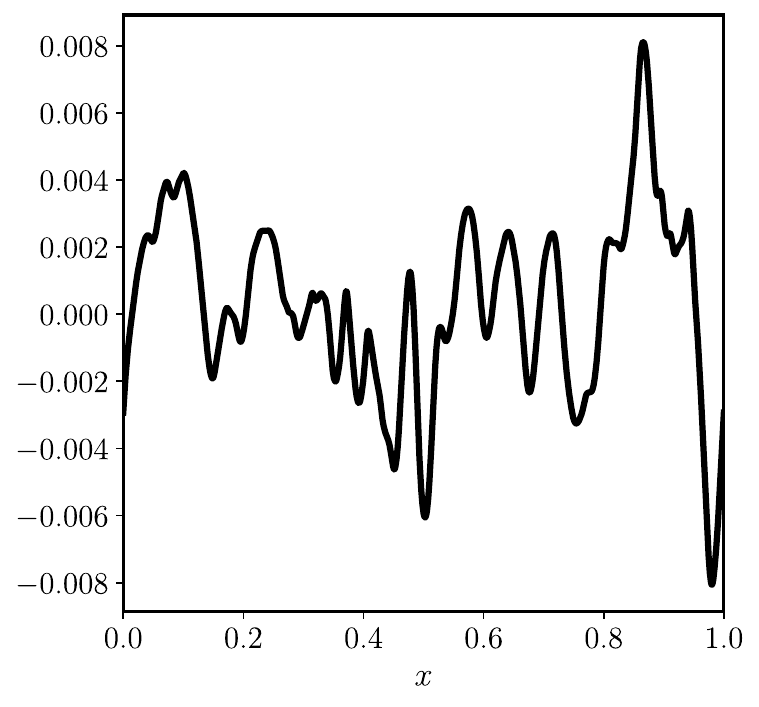}
                \caption{Pointwise error}
                \label{fig:pwerror_onesample}
        \end{subfigure}
        \vspace{-5mm}
        \caption{Representative input-output test sample for the Burgers' equation solution map $ \Fd\defeq \Psi_{1} $: Figure~\ref{fig:sample_burg}{\rm{(\subref{fig:prediction_onesample})}} shows a sample input, output (truth), and trained RFM prediction (test), while Figure~\ref{fig:sample_burg}{\rm{(\subref{fig:pwerror_onesample})}} displays the pointwise error. The relative $ L^2 $ error for this single prediction is $ 0.0146 $. Here, $ n=512 $, $ m=1024 $, and $ K=1025 $.}
        \label{fig:sample_burg}
\end{figure}

\begin{table}[tb]
\footnotesize
    \centering
    \caption{Expected relative error $ e_{n',m} $ for time upscaling with the learned RFM operator semigroup for Burgers' equation. Here, $ n'=4000 $, $ m=1024 $, $ n=512 $, and $ K=129 $. The RFM is trained on data from the evolution operator $ \Psi_{T=0.5}$ and then tested on input-output samples generated from $ \Psi_{jT} $, where $ j=2,\, 3,\, 4 $, by repeated composition of the learned model. The increase in error is small even after three compositions, reflecting excellent out-of-distribution performance.}
    \label{tab:time_upscale}
    \begin{tabular}{ @{}lccccc@{} }
            \toprule
            Train on: & $ T=0.5 $ & $\ \quad \ $ Test on: & $ 2T=1.0 $ & $ 3T=1.5 $ & $4T=2.0$ \\
            \midrule
            & 0.0360 & & 0.0407 & 0.0528 & 0.0788 \\
            \bottomrule
    \end{tabular}
\end{table}

Our experiments study the RFM approximation to the viscous Burgers' equation evolution operator semigroup~\cref{eqn:solnmap_burg}. As a visual aid for the high-dimensional problem at hand, \Cref{fig:sample_burg} shows a representative sample input and output along with a trained RFM test prediction. To determine whether the RFM has actually learned the correct evolution operator, we test the semigroup property of the map;~\cite{wu2020data} pursues closely related work also in a Fourier space setting. Denote the $ (j-1) $-fold composition of a function $ G $ with itself by $ G^{j} $. Then, with $ u(0,\cdot)=a $, we have
\begin{align}\label{eqn:semigroup}
(\Psi_{T}\circ \cdots \circ \Psi_{T})(a)=\Psi_{T}^{j}(a)=\Psi_{jT}(a)=u(jT, \cdot)
\end{align}
by definition. We train the RFM on input-output pairs from the map $ \Psi_{T} $ with $ T\defeq 0.5 $ to obtain $ \widehat{F}\defeq F_{m}(\slot;\widehat{\al}) $. Then, it should follow from~\cref{eqn:semigroup} that $ \widehat{F}^{j}\approx \Psi_{jT}$, that is, each application of $ \widehat{F} $ should evolve the solution $ T $ time units. We test this semigroup approximation by learning the map $ \widehat{F} $ and then comparing $ \widehat{F}^{j} $ on  $n'= 4000$ fixed inputs to outputs from each of the operators $ \Psi_{jT} $, with $ j\in\{1,2,3,4\} $ (the solutions at time $ T $, $ 2T $, $ 3T $, $ 4T $). The results are presented in~\Cref{tab:time_upscale} for a fixed mesh size $ K=129 $.
We observe that the composed RFM map $ \widehat{F}^{j} $ accurately captures $ \Psi_{jT} $, though this accuracy deteriorates as $ j $ increases due to error propagation in time as is common with any traditional integrator. However, even after three compositions corresponding to 1.5 time units past the training time $ T=0.5 $, the relative error only increases by around $ 0.04 $. It is remarkable that the RFM learns time evolution without explicitly time-stepping the PDE~\cref{eqn:burgers_ibvp} itself. Such a procedure is coined \emph{time upscaling} in the PDE context and in some sense breaks the CFL stability barrier~\cite{demanet2006curvelets}. \Cref{tab:time_upscale} is evidence that the RFM has excellent out-of-distribution performance: although only trained on inputs $ a\sim\nu $, the model outputs accurate predictions given new input
samples $ \Psi_{jT}(a) \sim (\Psi_{jT})_{\sharp}\nu $.

\begin{figure}[tb]
        \centering
        \begin{subfigure}[]{0.49\textwidth}
                \centering
                \includegraphics[width=\textwidth]{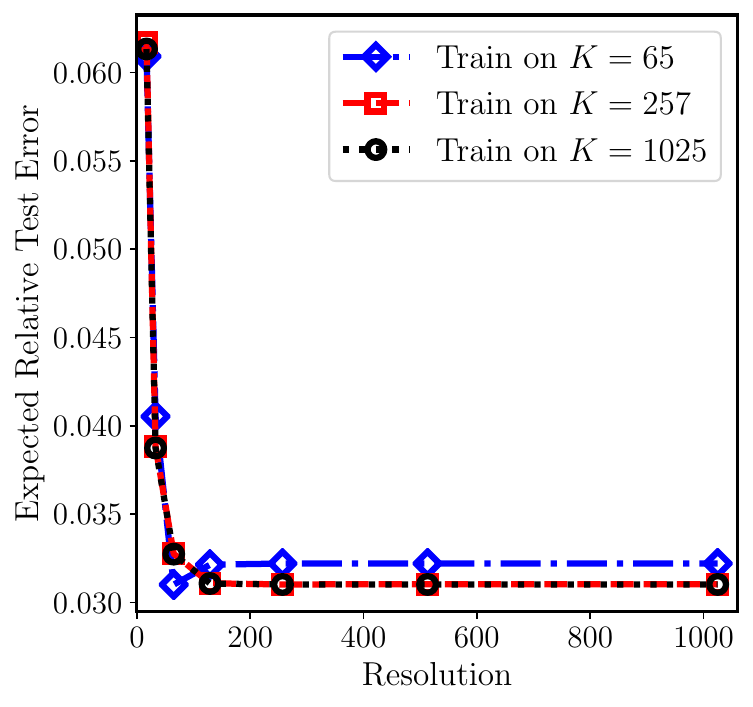}
                \caption{Test error vs. resolution}
                \label{fig:gridtransfer_burg_panel}
        \end{subfigure}%
        \hfill%
        \begin{subfigure}[]{0.48\textwidth}
                \centering
                \includegraphics[width=\textwidth]{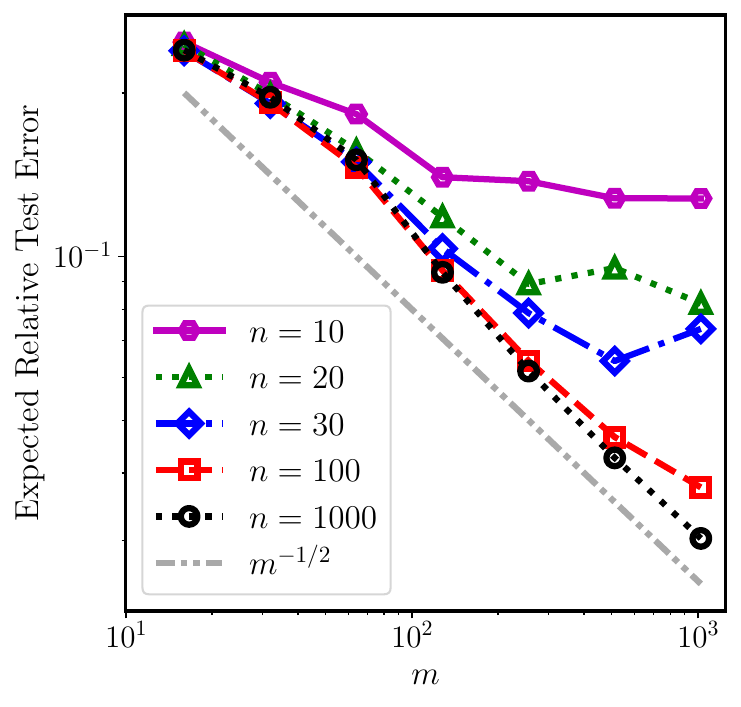}
                \caption{Test error vs. $m$ and $n$}
                \label{fig:gridsweep_burg_n}
        \end{subfigure}
        \vspace{-5mm}
        \caption{Expected relative test error of a trained RFM for the Burgers' evolution operator $ \Fd=\Psi_{1} $ with $ n'=4000 $ test pairs: Figure~\ref{fig:gridtranfser_burg}{\rm{(\subref{fig:gridtransfer_burg_panel})}} displays the invariance of test error w.r.t.~training and testing on different resolutions for $ m=1024 $ and $ n=512 $ fixed; the RFM can train and test on different mesh sizes without loss of accuracy. Figure~\ref{fig:gridtranfser_burg}{\rm{(\subref{fig:gridsweep_burg_n})}} shows the decay of the test error for resolution $ K=129 $ fixed as a function of $ m $ and $ n $; the error follows the $ O(m^{-1/2}) $ Monte Carlo rate remarkably well and the smallest error achieved is $ 0.0303 $ for $ n=1000 $ and $ m=1024 $.}
        \label{fig:gridtranfser_burg}
\end{figure}

We next study the ability of the RFM to transfer its learned coefficients $ \widehat{\al} $ obtained from training on mesh size $ K $ to different mesh resolutions $ K' $ in~Figure~\ref{fig:gridtranfser_burg}{\rm{(\subref{fig:gridtransfer_burg_panel})}}. We fix $ T\defeq 1 $ from here on and observe that the lowest test error occurs when $ K=K'$, that is, when the train and test resolutions are identical; this behavior was also observed in the contemporaneous work~\cite{li2020neural}. At very low resolutions, such as $ K=17 $ here, the test error is dominated by discretization error which can become quite large; for example, resolving conceptually infinite-dimensional objects such as the Fourier space--based feature map in \cref{eqn:rf_fourier} or the $ L^2 $ norms in \cref{eqn:test_error} with only $ 17 $ grid points gives bad accuracy. But outside this regime, the errors are essentially constant across resolution regardless of the training resolution $ K $, indicating that the RFM learns its optimal coefficients independently of the resolution and hence generalizes well to any desired mesh size. In fact,
the trained model could be deployed on different discretizations of the domain $ D $ (e.g., various choices of finite elements, graph-based/particle methods), not just with different mesh sizes. Practically speaking, this means that high resolution training sets can be subsampled to smaller mesh sizes $ K $ (yet still large enough to avoid large discretization error) for faster training, leading to a trained model with nearly the same accuracy at all higher resolutions.

The smallest expected relative test error achieved by the RFM is $ 0.0303 $ for the configuration in Figure~\ref{fig:gridtranfser_burg}{\rm{(\subref{fig:gridsweep_burg_n})}}. This excellent performance is encouraging because the error we report is of the same order of magnitude as that reported in \cite[sect.~5.1]{li2020fourier} for the same Burgers' solution operator that we study, but with slightly different problem parameter choices. We emphasize that the neural operator methods in that work are based on deep learning, which involves training NNs by solving a nonconvex optimization problem with stochastic gradient descent, while our random feature methods have orders of magnitude fewer trainable parameters that are easily optimized through convex optimization. 
In Figure~\ref{fig:gridtranfser_burg}{\rm{(\subref{fig:gridsweep_burg_n})}}, we see that for large enough $n$, the error empirically follows the $O(m^{-1/2})$ parameter complexity bound that is suggested by \cref{thm:rate}. This theorem does not directly apply here because it requires the regularization parameter $\lambda$ to be strictly positive and $\Fd$ to be in the RKHS of $(\varphi,\mu)$ from \cref{sec:burg_formulation}, which we do not verify. Nonetheless, Figure~\ref{fig:gridtranfser_burg}{\rm{(\subref{fig:gridsweep_burg_n})}} indicates that the error bounds for the trained RFM hold for a larger class of problems than the stated assumptions suggest.

\begin{figure}[tb]
        \centering
        \begin{subfigure}[]{0.49\textwidth}
                \centering
                \includegraphics[width=\textwidth]{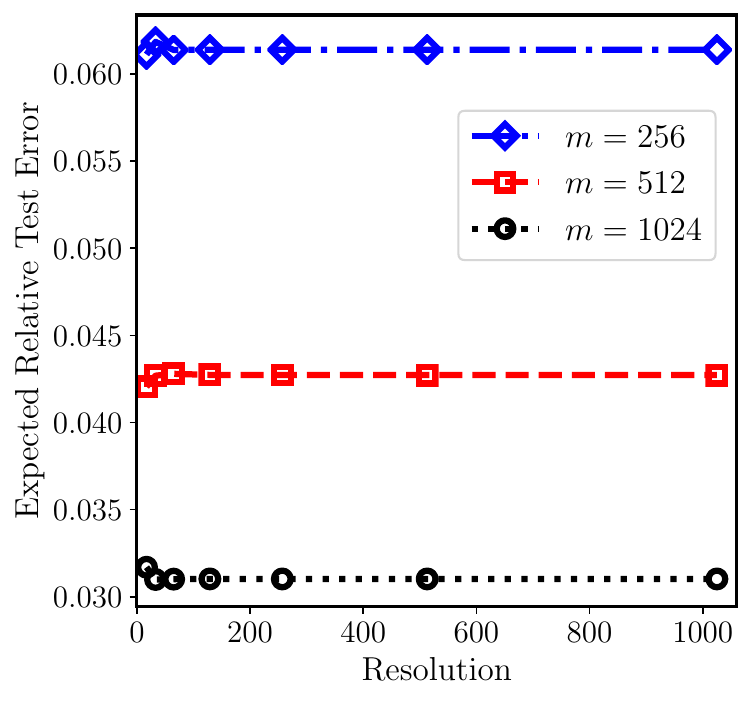}
                \caption{Test error vs. resolution}
                \label{fig:gridsweep_burg1}
        \end{subfigure}%
        \hfill%
        \begin{subfigure}[]{0.49\textwidth}
                \centering
                \includegraphics[width=\textwidth]{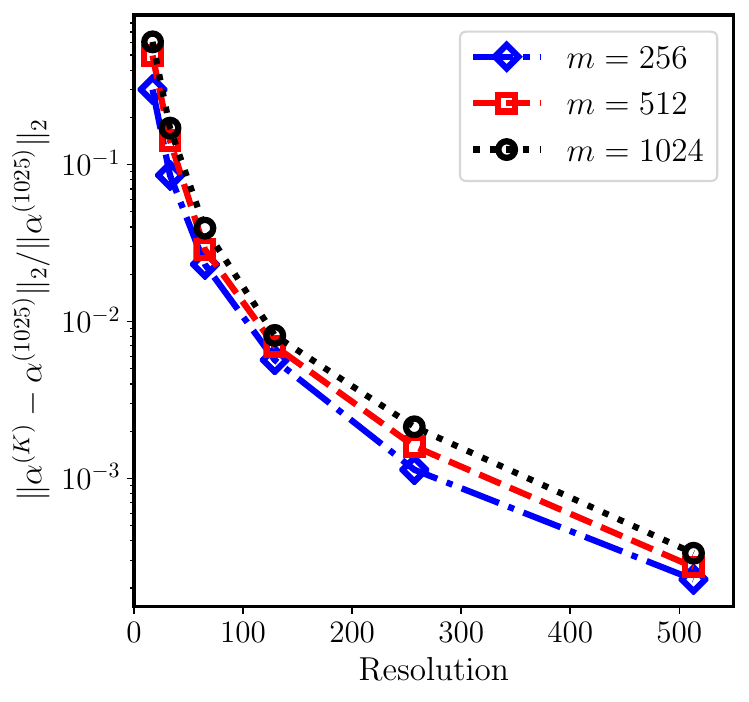}
                \caption{Minimizer error vs. resolution}
                \label{fig:gridsweep_burg2}
        \end{subfigure}
        \vspace{-5mm}
        \caption{Results of a trained RFM for the Burgers' equation evolution operator $ \Fd=\Psi_{1} $: Figure~\ref{fig:gridsweep_burg}{\rm{(\subref{fig:gridsweep_burg1})}} shows resolution-invariant test error for various $ m $. Figure~\ref{fig:gridsweep_burg}{\rm{(\subref{fig:gridsweep_burg2})}} displays the relative error of the learned coefficient $ \al $ w.r.t.~the coefficient learned on the highest mesh size ($ K=1025 $). Here, $ n=512 $ training and $ n'=4000 $ testing pairs were used.}
        \label{fig:gridsweep_burg}
\end{figure}

Finally, \Cref{fig:gridsweep_burg} demonstrates the invariance of the expected relative test error to the mesh resolution used for training and testing. This result is a consequence of framing the RFM on function space; other machine learning--based surrogate methods defined in finite dimensions exhibit an \emph{increase} in test error as mesh resolution is increased~(see \cite[sect.~4]{bhattacharya2020pca} for a numerical account of this phenomenon). Figure~\ref{fig:gridsweep_burg}{\rm{(\subref{fig:gridsweep_burg1})}} shows the error as a function of mesh resolution for three values of $ m $. For very low resolution, the error varies slightly but then flattens out to a constant value as $ K\to\infty $. Figure~\ref{fig:gridsweep_burg}{\rm{(\subref{fig:gridsweep_burg2})}} indicates that the learned coefficient $ \al^{(K)} $ for each $ K $ converges to some $ \al^{(\infty)} $ as $ K\to\infty $, again reflecting the design of the RFM as a mapping between infinite-dimensional spaces.

\subsection{Darcy Flow: Experiment}\label{sec:darcy_exp}
In this section, we consider Darcy flow on the physical domain $ D\defeq (0,1)^{2} $, the unit square. We generate a high resolution dataset of input-output pairs for $ \Fd $~\cref{eqn:solnmap_darcy} by solving~\cref{eqn:darcy_flow} on an equispaced $ 257\times 257 $ mesh (size $ K=257^{2} $) using a second order finite difference scheme. All mesh sizes $ K<257^{2} $ are subsampled from this original dataset and hence we consider numerical realizations of $ \Fd $ up to $ \R^{66049}\to\R^{66049} $. We denote \emph{resolution} by $ r $ such that $ K=r^{2} $. We fix $ n=128 $ training and $ n'=1000 $ testing pairs unless otherwise noted. The input data are drawn from the level set measure $ \nu $~\cref{eqn:prior_levelset} with $ \tau=3 $ and $ \al=2 $ fixed. We choose $ a^{+}=12 $ and $ a^{-}=3 $ in all experiments that follow and hence the contrast ratio $ a^{+}/a^{-}=4 $ is fixed. The source is fixed to $ f\equiv 1 $, the constant function. We evaluate the predictor-corrector random features $ \varphi $~\cref{eqn:rf_predictor_corrector} using an FFT-based fast Poisson solver corresponding to an underlying second order finite difference stencil at a cost of $ O(K\log K) $ per solve. The smoothed coefficient $ a_{\ep} $ in the definition of $ \varphi $ is obtained by solving~\cref{eqn:heat_mollify_neumann} with time step $ 0.03 $ and diffusion constant $ \eta=10^{-4} $; with centered second order finite differences, this incurs 34 time steps and hence a cost $O(34K)$. We fix the hyperparameters $ \al'=2 $, $ \tau'=7.5 $, $ s^{+}=1/12 $, $ s^{-}=-1/3 $, and $ \delta=0.15 $ for the map $ \varphi $. Unlike in~\cref{sec:burg_exp}, we find via grid search on $ \lambda $ that regularization during training does improve the reconstruction of the Darcy flow solution operator and hence we train with $ \lambda\defeq 10^{-8} $ fixed. We remark that, for simplicity, the above hyperparameters were not systematically and jointly optimized; as a consequence the RFM performance has the capacity to improve beyond the results in this section.

\begin{figure}[tb]
        \centering
        \begin{subfigure}[]{0.49\textwidth}
                \centering
                \includegraphics[width=\textwidth]{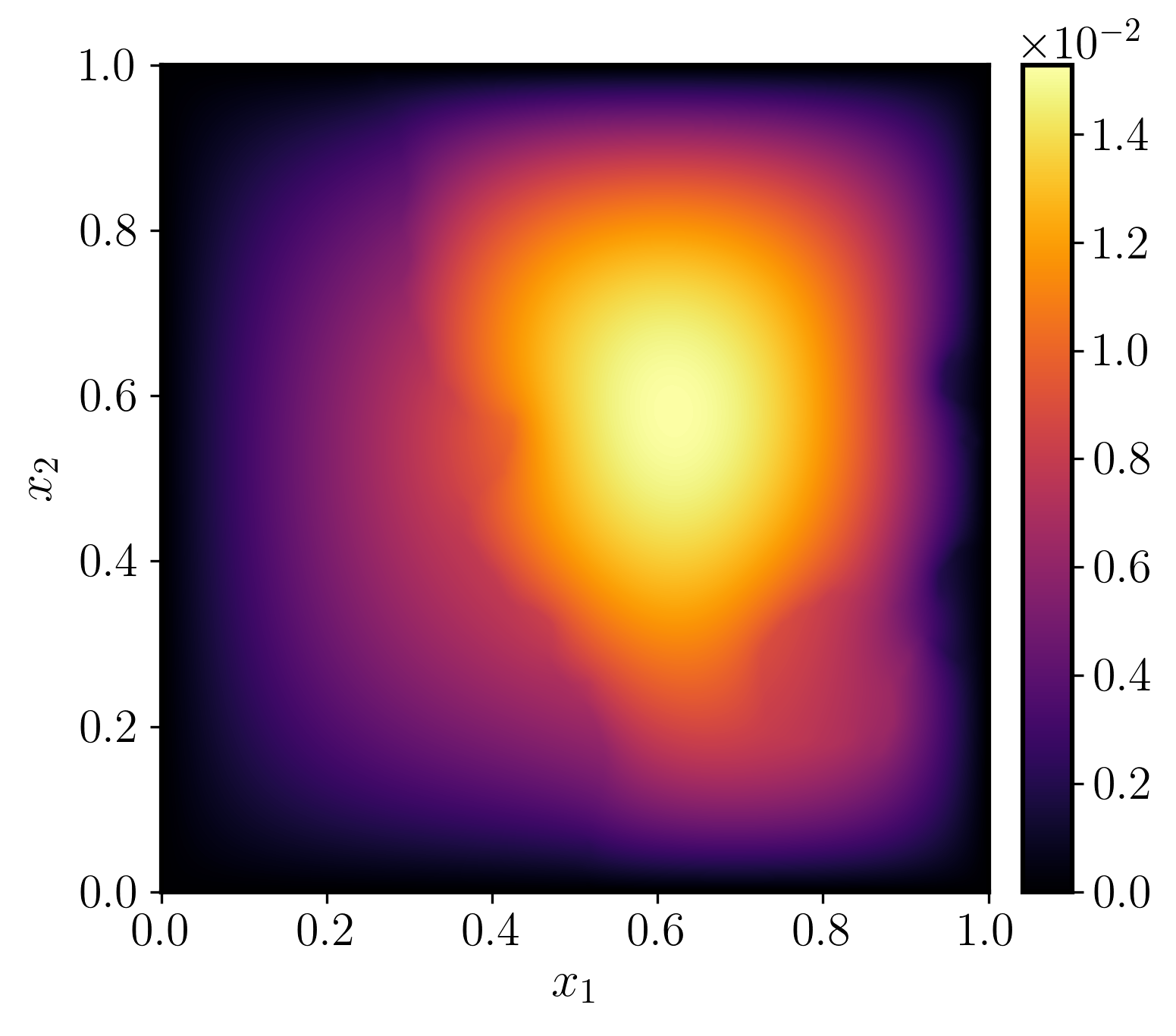}
                \caption{Truth}
                \label{fig:darcy_truth}
        \end{subfigure}%
        \hfill%
        \begin{subfigure}[]{0.49\textwidth}
                \centering
                \includegraphics[width=\textwidth]{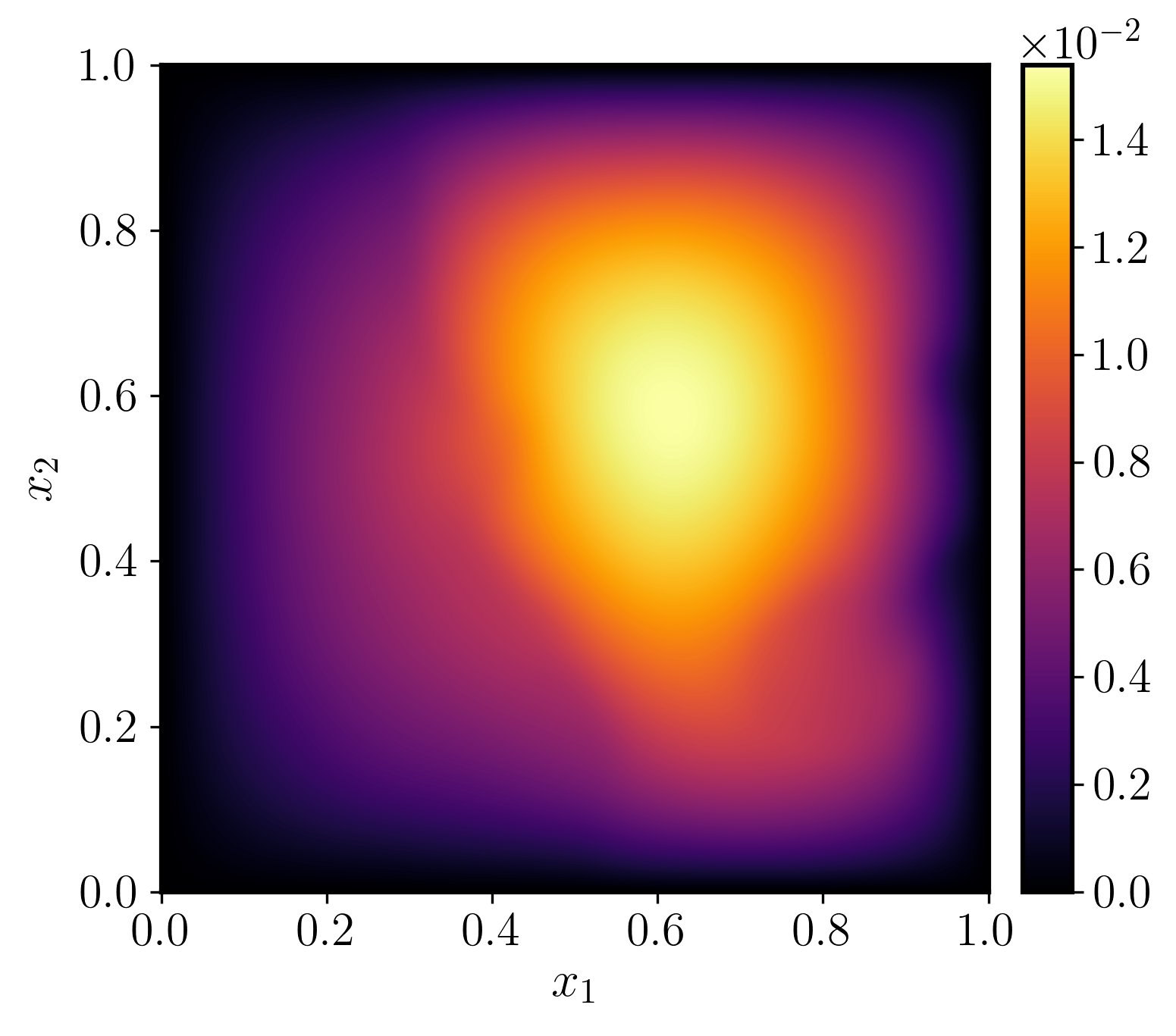}
                \caption{Approximation}
                \label{fig:darcy_predict}
        \end{subfigure}
        \begin{subfigure}[]{0.49\textwidth}
                \centering
                \includegraphics[width=\textwidth]{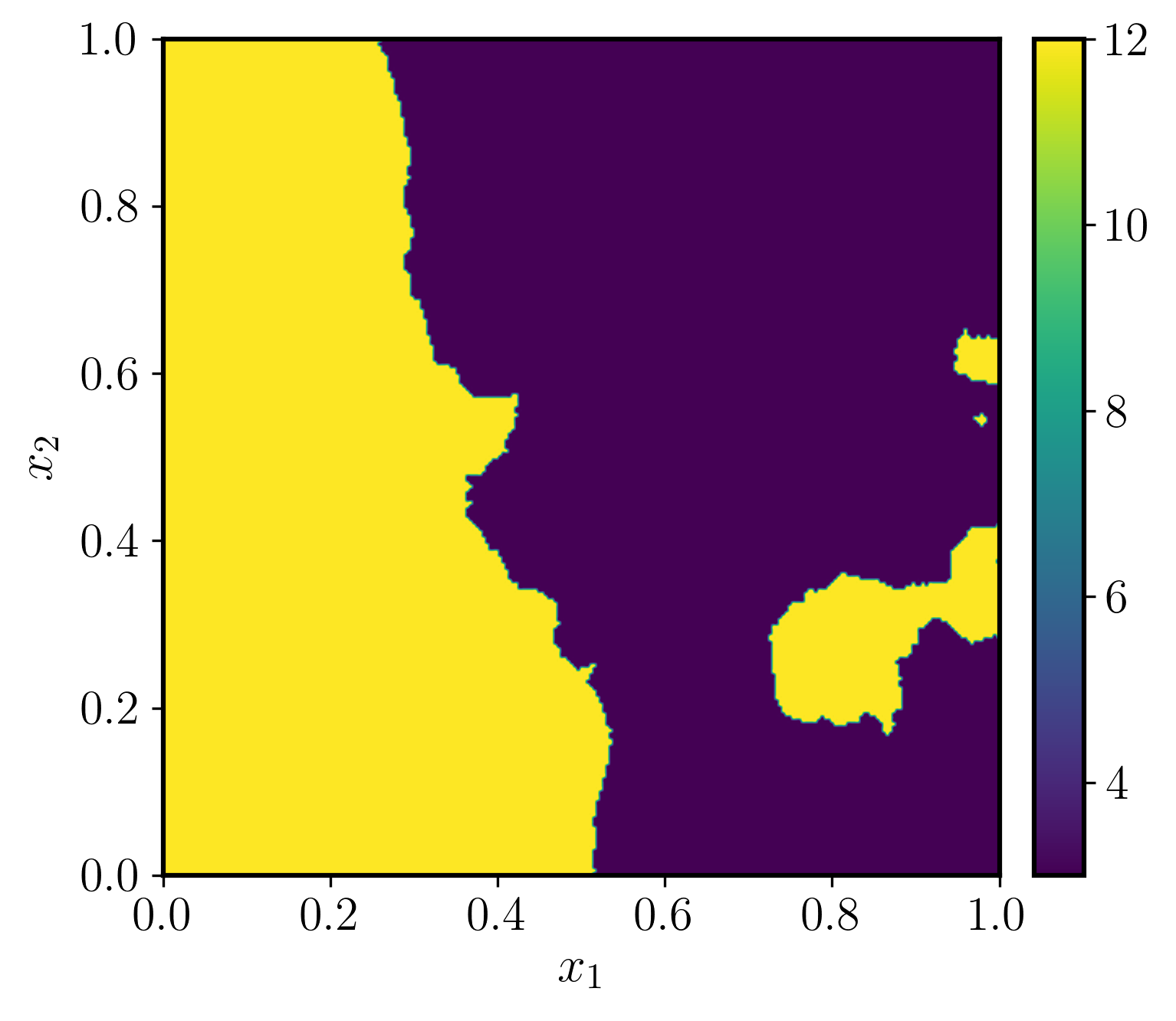}
                \caption{Input}
                \label{fig:darcy_input}
        \end{subfigure}%
        \hfill%
        \begin{subfigure}[]{0.49\textwidth}
                \centering
                \includegraphics[width=\textwidth]{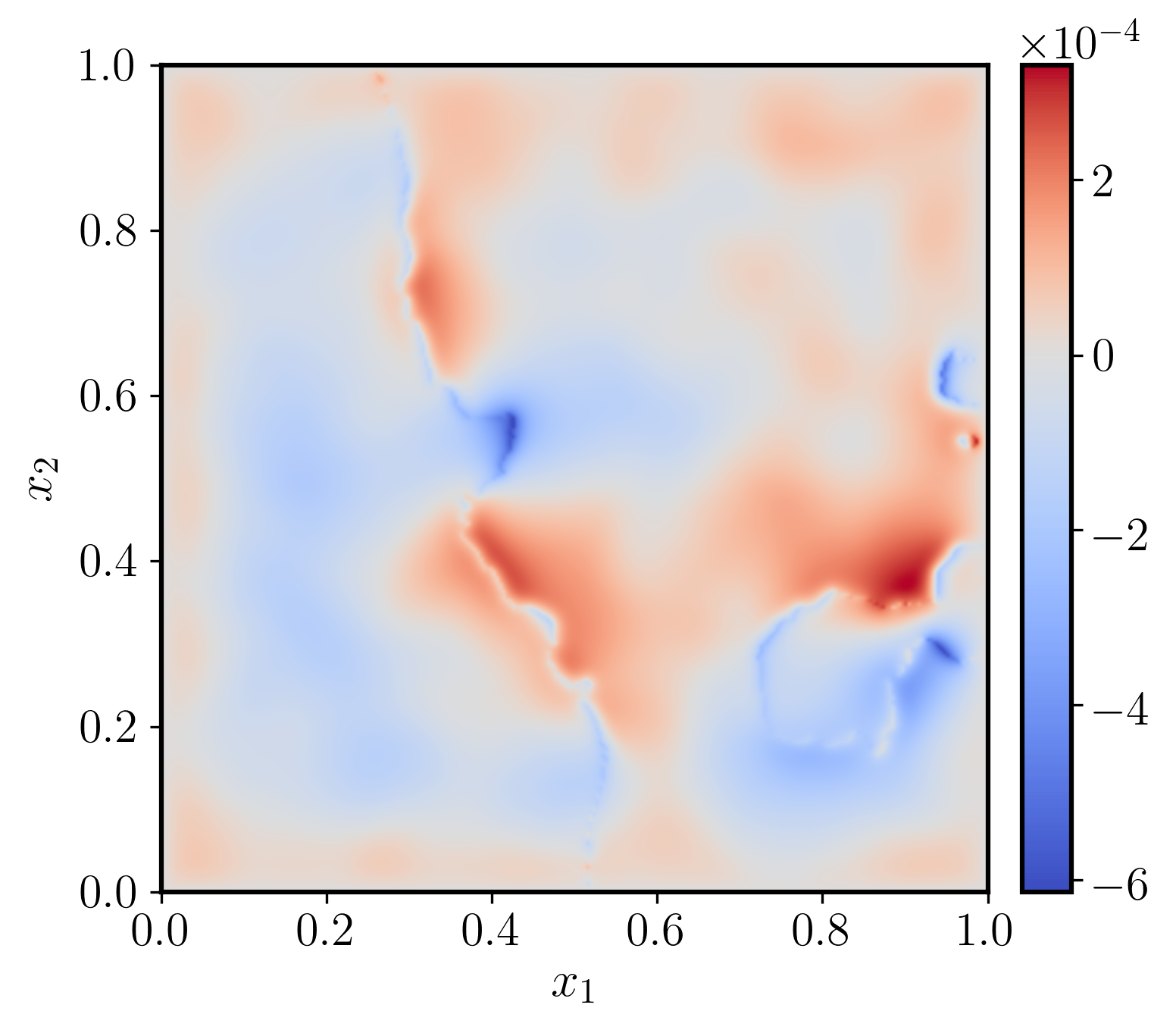}
                \caption{Pointwise error}
                \label{fig:darcy_pwerror}
        \end{subfigure}
        \vspace{-5mm}
        \caption{Representative input-output test sample for the Darcy flow solution map: Figure~\ref{fig:darcy_inputoutput}{\rm{(\subref{fig:darcy_input})}} shows a sample input, Figure~\ref{fig:darcy_inputoutput}{\rm{(\subref{fig:darcy_truth})}} the resulting output (truth), Figure~\ref{fig:darcy_inputoutput}{\rm{(\subref{fig:darcy_predict})}} a trained RFM prediction, and Figure~\ref{fig:darcy_inputoutput}{\rm{(\subref{fig:darcy_pwerror})}} the pointwise error. The relative $ L^2 $ error for this single prediction is $ 0.0122 $. Here, $ n=256 $, $ m=350 $, and $ K=257^{2} $.}
        \label{fig:darcy_inputoutput}
\end{figure}

\begin{figure}[tb]
        \centering
        \begin{subfigure}[]{0.49\textwidth}
                \centering
                \includegraphics[width=\textwidth]{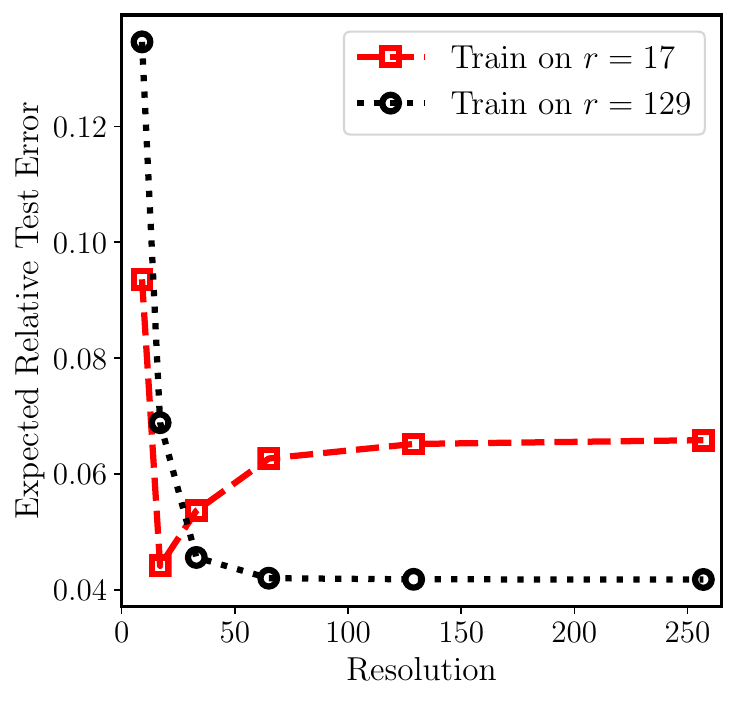}
                \caption{Test error vs. resolution}
                \label{fig:gridtransfer_darcy_panel}
        \end{subfigure}%
        \hfill%
        \begin{subfigure}[]{0.49\textwidth}
                \centering
                \includegraphics[width=\textwidth]{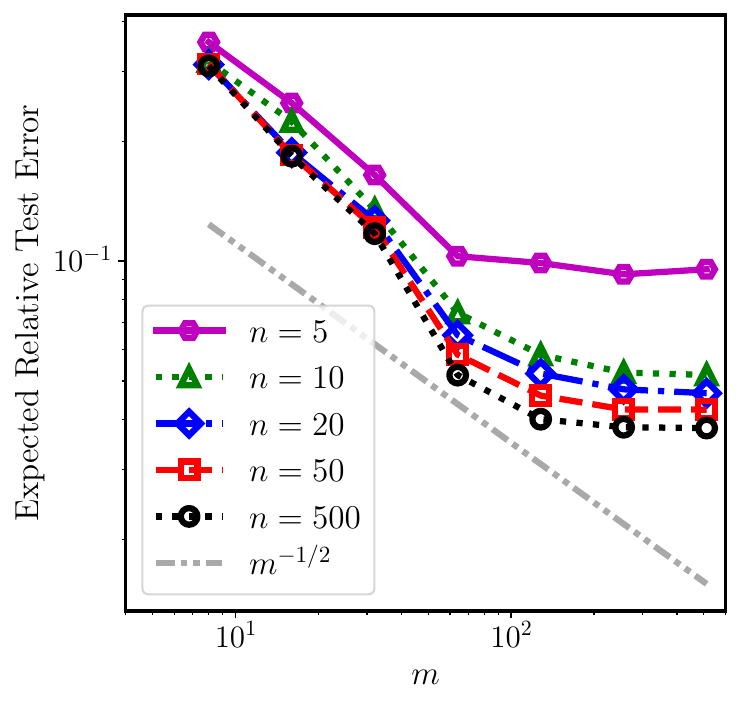}
                \caption{Test error vs. $m$ and $n$}
                \label{fig:gridsweep_darcy_n}
        \end{subfigure}
        \vspace{-5mm}
        \caption{Expected relative test error of a trained RFM for Darcy flow with $ n'= 1000 $ test pairs: Figure~\ref{fig:gridtranfser_darcy}{\rm{(\subref{fig:gridtransfer_darcy_panel})}} displays the invariance of test error w.r.t.~training and testing on different resolutions for $ m=512 $ and $ n=256 $ fixed; the RFM can train and test on different mesh sizes without significant loss of accuracy. Figure~\ref{fig:gridtranfser_darcy}{\rm{(\subref{fig:gridsweep_darcy_n})}} shows the decay of the test error for resolution $ r=33 $ fixed as a function of $ m $ and $ n $; the smallest error achieved is $ 0.0381 $ for $ n=500 $ and $ m=512 $.}
        \label{fig:gridtranfser_darcy}
\end{figure}

Darcy flow is characterized by the geometry of the high contrast coefficients $ a\sim\nu $. As seen in \Cref{fig:darcy_inputoutput}, the solution inherits the steep interfaces of the input. However, we see that a trained RFM with predictor-corrector random features \cref{eqn:rf_predictor_corrector} captures these interfaces well, albeit with slight smoothing; the error concentrates on the location of the interface. The effect of increasing $ m $ and $ n $ on the test error is shown in~Figure~\ref{fig:gridtranfser_darcy}{\rm{(\subref{fig:gridsweep_darcy_n})}}. Here, the error appears to saturate more than was observed for the Burgers' equation problem~(Figure~\ref{fig:gridtranfser_burg}{\rm{(\subref{fig:gridsweep_burg_n})}}) and does not follow the $O(m^{-1/2})$ rate. This is likely due to fixing $\lambda$ to be constant instead of scaling it with $m$ as suggested by \cref{thm:rate}. It is also possible that the Darcy flow solution map does not belong to the RKHS $ \cH_{k_{\mu}} $, leading to an additional misspecification error.
However, the smallest test error achieved for the best performing RFM configuration is $ 0.0381 $, which is on the same scale as the error reported in competing neural operator-based methods~\cite{bhattacharya2020pca,li2020neural} for the same setup.

The RFM is able to be successfully trained and tested on different resolutions for Darcy flow. Figure~\ref{fig:gridtranfser_darcy}{\rm{(\subref{fig:gridtransfer_darcy_panel})}} shows that, again, for low resolutions, the smallest relative test error is achieved when the train and test resolutions are identical (here, for $ r=17 $). However, when the resolution is increased away from this low resolution regime, the relative test error slightly increases then approaches a constant value, reflecting the function space design of the method. Training the RFM on a high resolution mesh poses no issues when transferring to lower or higher resolutions for model evaluation, and it achieves consistent error for test resolutions sufficiently large~(i.e., $ r \geq 33 $, the regime where discretization error starts to become negligible). Additionally, the RFM basis functions $ \{\varphi(\slot;\theta_j)\}_{j=1}^{m} $ are defined without any dependence on the training data unlike in other competing approaches based on similar shallow linear approximations, such as the reduced basis method or the PCA-Net method in~\cite{bhattacharya2020pca}. Consequently, our RFM may be directly evaluated on any desired mesh resolution once trained (``superresolution''), whereas those aforementioned approaches require some form of interpolation to transfer between different mesh sizes (see \cite[sect.~4.3]{bhattacharya2020pca}).

\begin{figure}[tb]
        \centering
        \begin{subfigure}[]{0.49\textwidth}
                \centering
                \includegraphics[width=\textwidth]{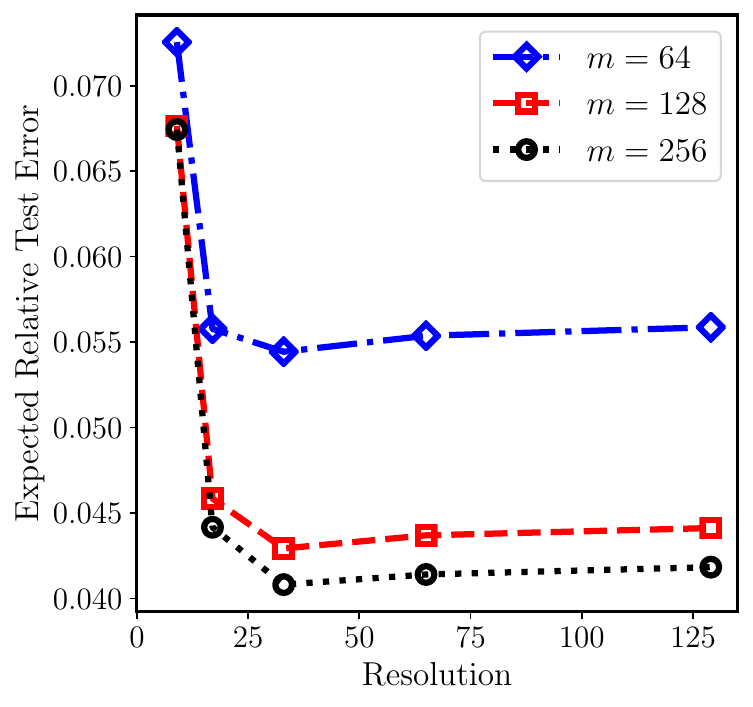}
                \caption{Test error vs. resolution}
                \label{fig:gridsweep_darcy1}
        \end{subfigure}%
        \hfill%
        \begin{subfigure}[]{0.49\textwidth}
                \centering
                \includegraphics[width=\textwidth]{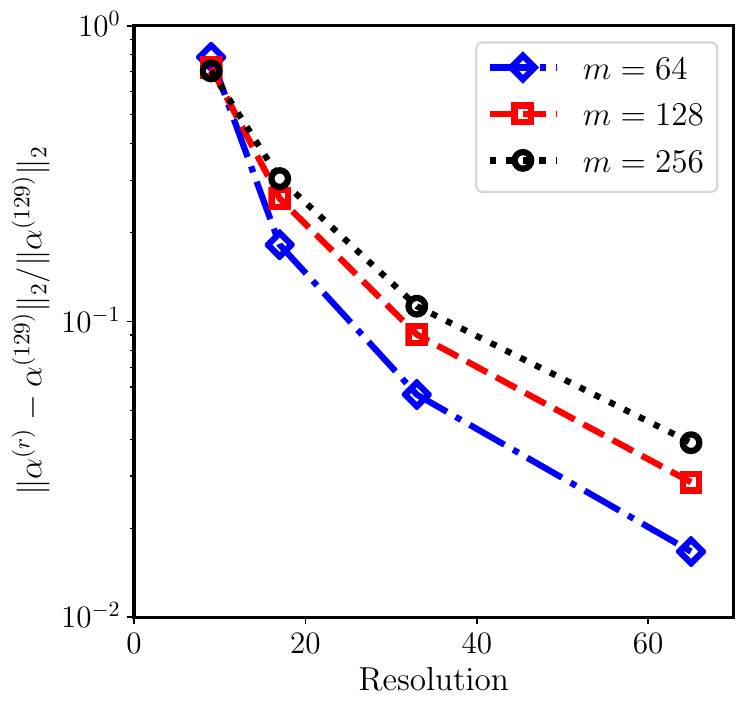}
                \caption{Minimizer error vs. resolution}
                \label{fig:gridsweep_darcy2}
        \end{subfigure}
        \vspace{-5mm}
        \caption{Results of a trained RFM for Darcy flow: Figure~\ref{fig:gridsweep_darcy}{\rm{(\subref{fig:gridsweep_darcy1})}} demonstrates resolution-invariant test error for various $ m $, while Figure~\ref{fig:gridsweep_darcy}{\rm{(\subref{fig:gridsweep_darcy2})}} displays the relative error of the learned coefficient $ \al^{(r)} $ at resolution $ r $ w.r.t.~the coefficient learned on the highest resolution ($ r=129 $). Here, $ n=128 $ training and $ n'=1000 $ testing pairs were used.}
        \label{fig:gridsweep_darcy}
\end{figure}

In~\Cref{fig:gridsweep_darcy}, we again confirm that our method is invariant to the refinement of the mesh and improves with more random features. While the difference at low resolutions is more pronounced than that observed for Burgers' equation, our results for Darcy flow still suggest that the expected relative test error converges to a constant value as resolution increases; an estimate of this rate of convergence is seen in~Figure~\ref{fig:gridsweep_darcy}{\rm{(\subref{fig:gridsweep_darcy2})}}, where we plot the relative error of the learned parameter $ \al^{(r)} $ at resolution $ r $ w.r.t.~the parameter learned at the highest resolution trained, which was $ r=129 $.

\section{Conclusion}
\label{sec:conclusion}
This paper introduces a random feature methodology for the data-driven estimation of operators mapping between infinite-dimensional Banach spaces. It may be interpreted as a low-rank approximation to operator-valued kernel ridge regression. Training the function-valued random features only requires solving a quadratic optimization problem for an $ m $-dimensional coefficient vector.
The conceptually infinite-dimensional algorithm is nonintrusive and
results in a scalable method that is consistent with the continuum limit, robust to discretization, and highly flexible in practical use cases. Numerical experiments confirm these benefits in scientific machine learning applications involving two nonlinear forward operators arising from PDEs. Backed by tractable training routines and theoretical guarantees, operator learning with the function-valued random features method displays considerable potential for accelerating many-query computational tasks and for discovering new models from high-dimensional experimental data in science and engineering.

Going beyond this paper, several directions for future research remain open.
Some of the first theoretical results for function-valued random features are summarized in \cref{sec:theory}. However, it is not yet known what conditions on the problem and the feature pair allow for faster rates of convergence. In addition, it is of interest to characterize the quality of the operator RKHS spaces induced by random feature pairs and whether practical problem classes actually belong to these spaces. Also of importance is the question of how to automatically adapt function-valued random features to data instead of manually constructing them. Some possibilities along this line of work include the Bayesian estimation of hyperparameters, as is frequently used in Gaussian process regression, or more general hierarchical learning of the random feature pair $ (\varphi, \mu) $ itself; the work \cite{dunbar2024hyperparameter} contains preliminary results of this type. In tandem, there is a need for a mature function-valued random features software library that includes efficient linear solvers and GPU implementations, benchmark problems, and robust hyperparameter optimizers. These advances will further enable the random features method to learn from real-world data and solve challenging forward and inverse problems from the physical sciences, such as climate modeling and material modeling, with controlled computational complexity.

\section*{Data and Code Availability}
Links to datasets and code used to produce the numerical results and figures in this paper are available at
\begin{center}
    \url{https://github.com/nickhnelsen/random-features-banach}\,.
\end{center}

\section*{Acknowledgments}
The authors thank Bamdad Hosseini and Nikola Kovachki for helpful discussions and are grateful to the two anonymous referees for their careful reading of and insightful comments on the original publication of this paper. They are also grateful to Samuel Lanthaler for reading and commenting on an earlier draft of the present paper.

\end{document}